\documentclass[10pt,twocolumn,letterpaper]{article}

\usepackage{color}
\usepackage{xcolor}

\usepackage{iccv}
\usepackage{bbding}
\usepackage{times}
\usepackage{epsfig}
\usepackage{graphicx}
\usepackage{amsmath}
\usepackage{amssymb}
\usepackage{subfigure}
\usepackage{multirow}
\usepackage{adjustbox}
\usepackage{comment}
\usepackage{boldline}
\usepackage{amssymb}
\usepackage{algorithmic}
\usepackage{float}

\usepackage{fullpage}
\usepackage{fancyhdr}
\usepackage[ruled,vlined]{algorithm2e}

\usepackage{tikz}
\newcommand{\tikzxmark}{%
\tikz[scale=0.23] {
    \draw[line width=0.7,line cap=round] (0,0) to [bend left=6] (1,1);
    \draw[line width=0.7,line cap=round] (0.2,0.95) to [bend right=3] (0.8,0.05);
}}
\newcommand{\tikzcmark}{%
\tikz[scale=0.23] {
    \draw[line width=0.7,line cap=round] (0.25,0) to [bend left=10] (1,1);
    \draw[line width=0.8,line cap=round] (0,0.35) to [bend right=1] (0.23,0);
}}

\usepackage[colorlinks=true, breaklinks=true,bookmarks=false]{hyperref}

\iccvfinalcopy 

\def\iccvPaperID{****} 

\ificcvfinal\fi

\begin{document}

\title{GraphEcho: Graph-Driven Unsupervised Domain Adaptation for Echocardiogram Video Segmentation}

\author{Jiewen Yang\textsuperscript{1} ~ Xinpeng Ding\textsuperscript{1} ~ Ziyang Zheng\textsuperscript{1$\dag$} ~ Xiaowei Xu\textsuperscript{2}$^*$ ~ Xiaomeng Li\textsuperscript{1}$^*$ \\
The Hong Kong University of Science and Technology\textsuperscript{1} \\
Guangdong Provincial People's Hospital, Institute of Cardiovascular Diseases, GuangZhou, China\textsuperscript{2}\\
{\tt\small \{jyangcu, xdingaf\}@connect.ust.hk, xiao.Wei.xu@foxmail.com, eexmli@ust.hk}
}
\maketitle
\ificcvfinal\thispagestyle{empty}\fi

\def\thefootnote{*}\footnotetext{Corresponding author}\def\thefootnote{\arabic{footnote}}
\def\thefootnote{$\dag$}\footnotetext{Interning at The Hong Kong University of Science and Technology}\def\thefootnote{\arabic{footnote}}

\begin{abstract}
  Echocardiogram video segmentation plays an important role in cardiac disease diagnosis. This paper studies the unsupervised domain adaption (UDA) for echocardiogram video segmentation, where the goal is to generalize the model trained on the source domain to other unlabelled target domains. 
  Existing UDA segmentation methods are not suitable for this task because they do not model local information and the cyclical consistency of heartbeat.
  In this paper, we introduce a newly collected \textbf{CardiacUDA dataset} and a novel \textbf{GraphEcho} method 
  for cardiac structure segmentation. 
  Our GraphEcho comprises two innovative modules, the Spatial-wise Cross-domain Graph Matching (SCGM) and the Temporal Cycle Consistency (TCC) module, which utilize prior knowledge of echocardiogram videos, i.e., consistent cardiac structure across patients and centers and the heartbeat cyclical consistency, respectively. These two modules can better align global and local features from source and target domains, leading to improved UDA segmentation results. Experimental results showed that our GraphEcho outperforms existing state-of-the-art UDA segmentation methods. Our collected dataset and code will be publicly released upon acceptance. This work will lay a new and solid cornerstone for cardiac structure segmentation from echocardiogram videos. Code and dataset are available at :~\href{https://github.com/xmed-lab/GraphEcho}{https://github.com/xmed-lab/GraphEcho}


\end{abstract}

  
\vspace{-10pt}
\section{Introduction}
\label{sec:intro}


%
  \begin{figure}[t]
    \begin{center}
    \includegraphics[width=0.999\linewidth]{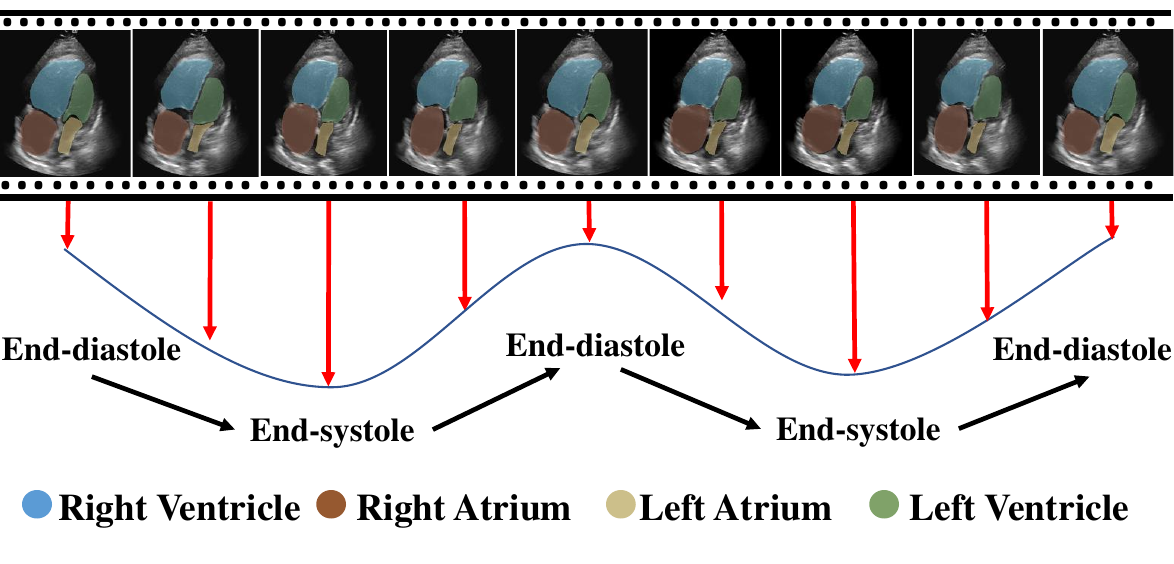}
    \caption{Examples of nine frames from our newly collected \textbf{CardiacUDA} dataset, which serves as a new domain adaptation benchmark for cardiac structure segmentation from echocardiogram videos.}
    \vspace{-25pt}
    \end{center}
    \label{Fig:motivation}
  \end{figure}

\begin{figure}[!t]
    \centering
    \subfigure[CAMUS]{\label{fig:a}\includegraphics[width=19mm, height=19mm]{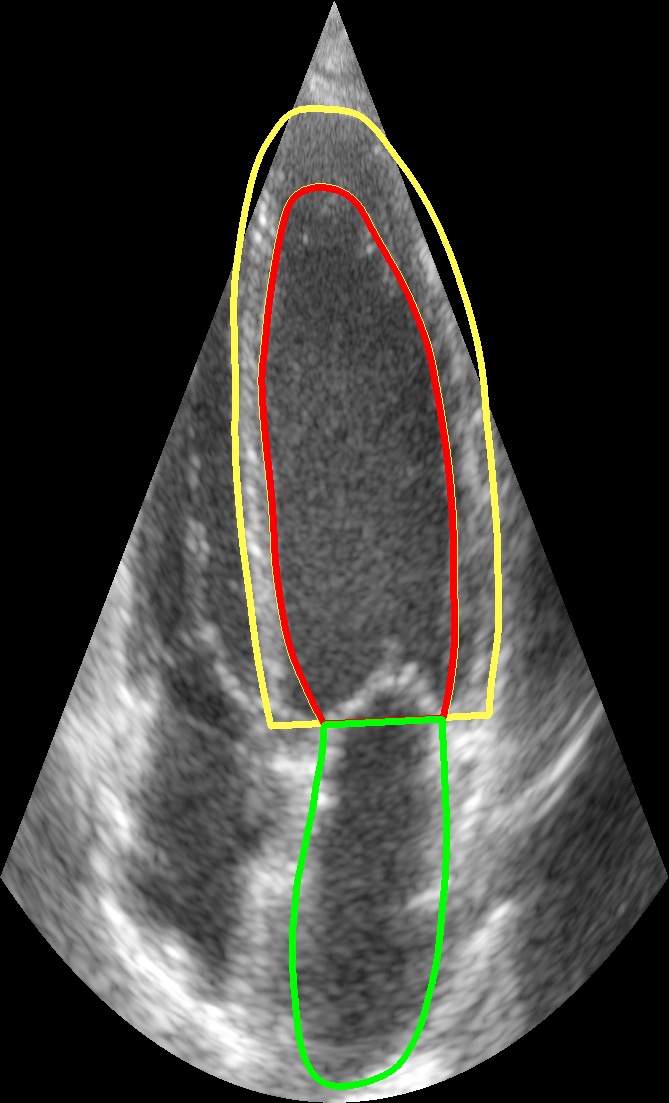}}
    \subfigure[Echonet]{\label{fig:b}\includegraphics[width=19mm]{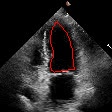}}
    \subfigure[Site G]{\label{fig:c}\includegraphics[width=19mm]{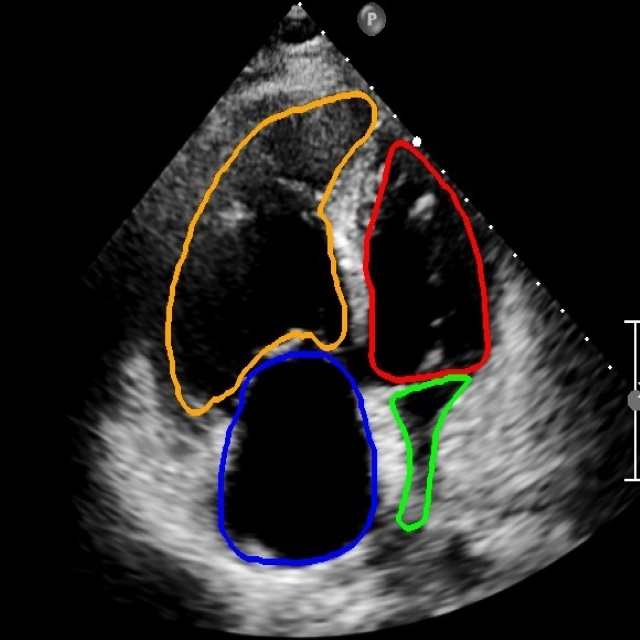}}
    \subfigure[Site R]{\label{fig:d}\includegraphics[width=19mm]{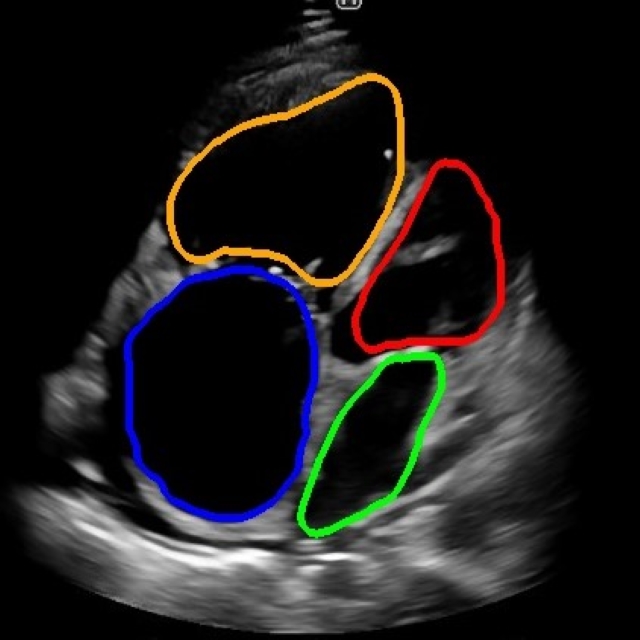}}
    \caption{(a-b) two public datasets; (c-d) our newly collected CardiacUDA from two sites: G and R. Red, green, blue, orange and yellow refer to the segmentation contours for left ventricle (LV), left atrium (RA), right Atrium (RA), right ventricle (RV), and epicardium of left ventricle, respectively. Table~\ref{tab:dataset_comp} outlines the advantages of our dataset over CAMUS and Echonet.}
    \vspace{-5pt}
    \label{fig:different datasets}
  \end{figure}

Echocardiography is a non-invasive diagnostic tool that enables the observation of all the structures of the heart. It can capture dynamic information on cardiac motion and function~\cite{papolos2016us,douglas2011accf,hughes2021deep}, making it a safe and cost-effective option for cardiac morphological and functional analysis. 
Accurate segmentation of cardiac structure, such as Left Ventricle (LV), Right Ventricle (RV), Left Atrium (LA), and Right Atrium (RA), is crucial for determining essential cardiac functional parameters, such as ejection fraction and myocardial strain. These parameters can assist physicians in identifying heart diseases,  planning treatments, and monitoring progress~\cite{fletcher2021machine,ouyang2020video}. 
Therefore, the development of an automated structure segmentation method for echocardiogram videos is of great significance.
Nonetheless, a model trained using data obtained from a specific medical institution may not perform as effectively on data collected from other institutions. For example, when a model trained on site G is directly tested on site R, its performance can significantly decrease to 48.5\% Dice, which is significantly lower than the performance of a model trained directly on site R, which achieves 81.3\% Dice; see results of  \textit{Without DA} and \textit{Upper Bound} in Table~\ref{tab:Result_GY_RMYY}. 
The result indicates that there are clear domain gaps between echocardiogram videos collected on different sites; see (c-d) in Figure~\ref{fig:different datasets}. Therefore, it is highly desirable to develop an unsupervised domain adaptation (UDA) method for cardiac structure segmentation from echocardiogram videos.


%
To the best of our knowledge, the UDA segmentation for echocardiogram videos has not been explored yet, and the most intuitive way is to adapt existing UDA methods designed for natural image segmentation and medical image segmentation to our task.
In general, existing methods can be grouped into 1). the image-level alignment methods \cite{kang2020pixel,melas2021pixmatch,yang2020fda,wang2020classes} that focus on aligning the style difference to minimize the domain gaps, such as PLCA~\cite{kang2020pixel}, PixMatch~\cite{melas2021pixmatch} and Fourier-base UDA~\cite{yang2020fda,wang2020classes}; 2). feature-level alignment methods~\cite{huang2022category,li2020content,li2022class}, such as~\cite{li2020content}, use global class-wise alignment to reduce the discrepancy between source and target domains.
However, applying these methods directly to cardiac structure segmentation in echocardiogram videos generated unsatisfactory performance; see results in Table~\ref{tab:Result_GY_RMYY} and Figure~\ref{Fig:visulization}. We thus consider two possible reasons: 
(1) Existing UDA methods~\cite{li2020content,kang2020pixel,melas2021pixmatch,yang2020fda,wang2020classes} primarily focused on aligning the global representations between the source and target domain while neglecting local information, such as LV, RV, LA, and RA; see (c-d) in Figure~\ref{fig:different datasets}. The failure to model local  information during adaptation leads to  restricted cardiac structure segmentation results. 
(2) Most existing methods~\cite{guan2021domain,li2020content,kang2020pixel,melas2021pixmatch,yang2020fda,wang2020classes,wu2021unsupervised,yao2022enhancing,xie2022unsupervised} were mainly designed for 2D or 3D images, which does not consider the video sequences and the cyclic properties of the cardiac cycle in our task. 
Given that heartbeat is a periodically recurring process, it is essential to ensure that the extracted features exhibit cyclical consistency~\cite{dai2022cyclical}. 

\begin{table}[!t]
\centering
\caption{The comparison of CardiacUDA, CAMUS, and Echonet. $\dagger$: 5 frames are labelled for each video in the training set, and all frames are labelled for each video in the validation and test dataset. }
\resizebox{0.48\textwidth}{!}{\begin{tabular}{c|ccc}
\hlineB{3}
Dataset             & \textbf{\small{Our CardiacUDA}}           & CAMUS~\cite{leclerc2019deep}                & EchoNet~\cite{ouyang2020video}              \\ \hlineB{3}
Video Num.       & 992               &  500             &  10,030           \\
Frames Num.   & 102,796              &  10,000              &  1,755,250          \\
Train/Test Labels     & \multicolumn{1}{c}{5 frames / Full~$\dagger$} & \multicolumn{1}{c}{2 frames / 2 frames} & \multicolumn{1}{c}{2 frames / 2 frames } \\
Multiple Centers    & $\checkmark$            & $\times$      & $\times$                    \\
Cardiac Views        & 4                    & 1                    & 1                    \\
Resolution          & 720p                & 480p                 & 120p                \\  
Annotated Regions          & LV, RV, LA, RA                & LV, LA                 & LV                 \\ \hlineB{3}
\end{tabular}}
\vspace{-15pt}
\label{tab:dataset_comp}
\end{table}

To address the above limitations, we present a novel graph-driven UDA method, namely \textbf{GraphEcho}, for echocardiogram video segmentation. Our proposed GraphEcho consists of two novel designs: \textbf{\textit {(1) Spatial-wise Cross-domain Graph Matching (SCGM)}} module and \textbf{\textit {(2) Temporal Cycle Consistency (TCC)}} module.
SCGM is motivated by the fact that the structure/positions of the different cardiac structures are similar across different patients and domains. 
For example, the left ventricle's appearance is typically visually alike across different patients; see red contours in Figure~\ref{fig:different datasets}. 
Our SCGM approach reframes domain alignment as a fine-grained graph-matching process that aligns both class-specific representations (local information) and the relationships between different classes (global information). By doing so, we can simultaneously improve intra-class coherence and inter-class distinctiveness.



Our TCC module is inspired by the observation the recorded echocardiogram videos exhibit cyclical consistency; see examples in Figure~\ref{Fig:motivation}. 
Specifically, our TCC module utilizes a series of recursive graph convolutional cells to model the temporal relationships between graphs across frames, generating a global temporal graph representation for each patient. We then utilized a contrastive objective that brings together representations from the same video while pushing away those from different videos, thereby enhancing temporal discrimination.
By integrating SCGM and TCC, our proposed method can leverage prior knowledge in echocardiogram videos to enhance inter-class differences and intra-class similarities across source and target domains while preserving temporal cyclical consistency, leading to a better UDA segmentation result. 


In addition, we collect a new dataset, called \textbf{CardiacUDA} from two clinical centers. 
As shown in Table~\ref{tab:dataset_comp}, compared to existing publicly available echocardiogram video datasets~\cite{leclerc2019deep,ouyang2020video}, our new dataset has higher resolutions, greater numbers of annotations, more annotated structure types as well as more scanning views. Our contribution can be summarized as follows: 


\begin{itemize}
  \item We will publicly release a newly collected echocardiogram video dataset, which can serve as a new benchmark dataset for video-based cardiac structure segmentation.
  \item We propose GraphEcho for cardiac structure segmentation, which incorporates a novel SCGM module and a novel TCC module that are motivated by prior knowledge. These modules effectively enhance both inter-class differences and intra-class similarities while preserving temporal cyclical consistency, resulting in superior UDA results.
  \item GraphEcho achieved superior performance compared to state-of-the-art UDA methods in both the computer vision and medical image analysis domains.
\end{itemize}

\section{Related Work}

\subsection{UDA for Segmentation}
In this section, we review the existing UDA segmentation methods for natural and medical images separately.

\noindent\textbf{Natural image segmentation.} For natural image segmentation, the adversarial-based domain adaptation  methods~\cite{gong2019dlow,melas2021pixmatch,tsai2018learning,vu2019advent,zhu2021cross} and multi-stage self-training methods, including single stage~\cite{chen2020self,liu2021cycle,mei2020instance,xie2022towards,zou2019confidence} and multi-stage~\cite{li2020content,li2022class} are the most commonly used training methods. 
The adversarial method aims to align the distributions and reduce the discrepancy of source and target domains through the Generative Adversarial Networks (GAN)~\cite{goodfellow2020generative} framework. At the same time, the self-training generate and update pseudo label online during training, such as applying data augmentation or domain mix-up. 

\noindent\textbf{Medical image segmentation.} For medical image segmentation, the UDA segmentation methods can be classified into image-level~\cite{chen2020unsupervised,wu2021unsupervised} that use GANs and different types of data augmentation to transfer source domain data to the target domain, and feature-level methods~\cite{guan2021domain,xie2022unsupervised}, such as feature alignment methods that aim to learn domain-invariant features across domains. 

While existing methods tend to overlook the temporal consistency characteristics in heartbeat cycles and the local relationships between different chambers across domains, our proposed GraphEcho method effectively learns both inter-class differences and intra-class coherence while preserving temporal consistency. This leads to superior UDA segmentation results.

\subsection{Graph Neural Networks }
Graph neural networks (GNNs) have the ability to construct graphical representations to describe irregular objects of data~\cite{han2022vision}. Also, graphs can iteratively aggregate the knowledge based on the broadcasting of their neighbouring nodes in the graph, which is more flexible for constructing the relationship among different components~\cite{han2022vision}.
The learned graph representations can be used in various downstream tasks, such as classification~\cite{han2022vision}, object detection~\cite{li2022sigma}, vision-language~\cite{chen2020graph}, etc.
Specifically, ViG~\cite{han2022vision} models an image as a graph and uses GNN to extract high-level features for image classification.
Li~\etal~\cite{li2022sigma} apply graphical representation instead of the feature space to explore multiple long-range contextual patterns from the different scales for more accurate object detection. 
GOT~\cite{chen2020graph} leverages the graphs to conduct the vision and language alignment for image-text retrieval.
There also exist some works that use the graph to conduct cross-domain alignment for object detection~\cite{li2022sigma} and classification~\cite{ding2018graph,ma2019gcan}.
However, these methods only capture the global graph information for images, which is insufficient for video segmentation tasks.
In this paper, our proposed GraphEcho learns both local class-wise and temporal-wise graph representations, which can reduce the domain gap in a fine-grained approach and enhance temporal consistency, leading to an enhanced result. 

\if
Currently, there exist many graph-based methods for different tasks. GAKT~\cite{ding2018graph} and GCAN~\cite{ma2019gcan} adapt the graph-based manner for UDA in the recognition task by propagating graph-based labels and exploiting class-level structure across domains, respectively. SIGMA\cite{li2022sigma} align the class-conditional distribution with graph matching in object detection task and GOT\cite{chen2020graph} optimal transport for graph matching to tackle cross-domain alignment problems in text-visual tasks.

Despite these methods achieving high-quality results in multiple tasks, treating the entirety instance for domain adaptation is able to solve coarse-grained problems. Those methods have not addressed the problem that occurs in the domain adaptation of echocardiogram video segmentation task that exists distinctive features in both spatial and temporal. Our MorGraphEcho method proposes two improved graph-based modules that can be introduced as spatial and temporal solvers for domain adaptation that considers spatiotemporal information when adapting the graph for the fine-grained problem.
\fi

\section{Method}
\begin{figure*}[!t]
  \begin{center}
  \includegraphics[width=0.999\linewidth]{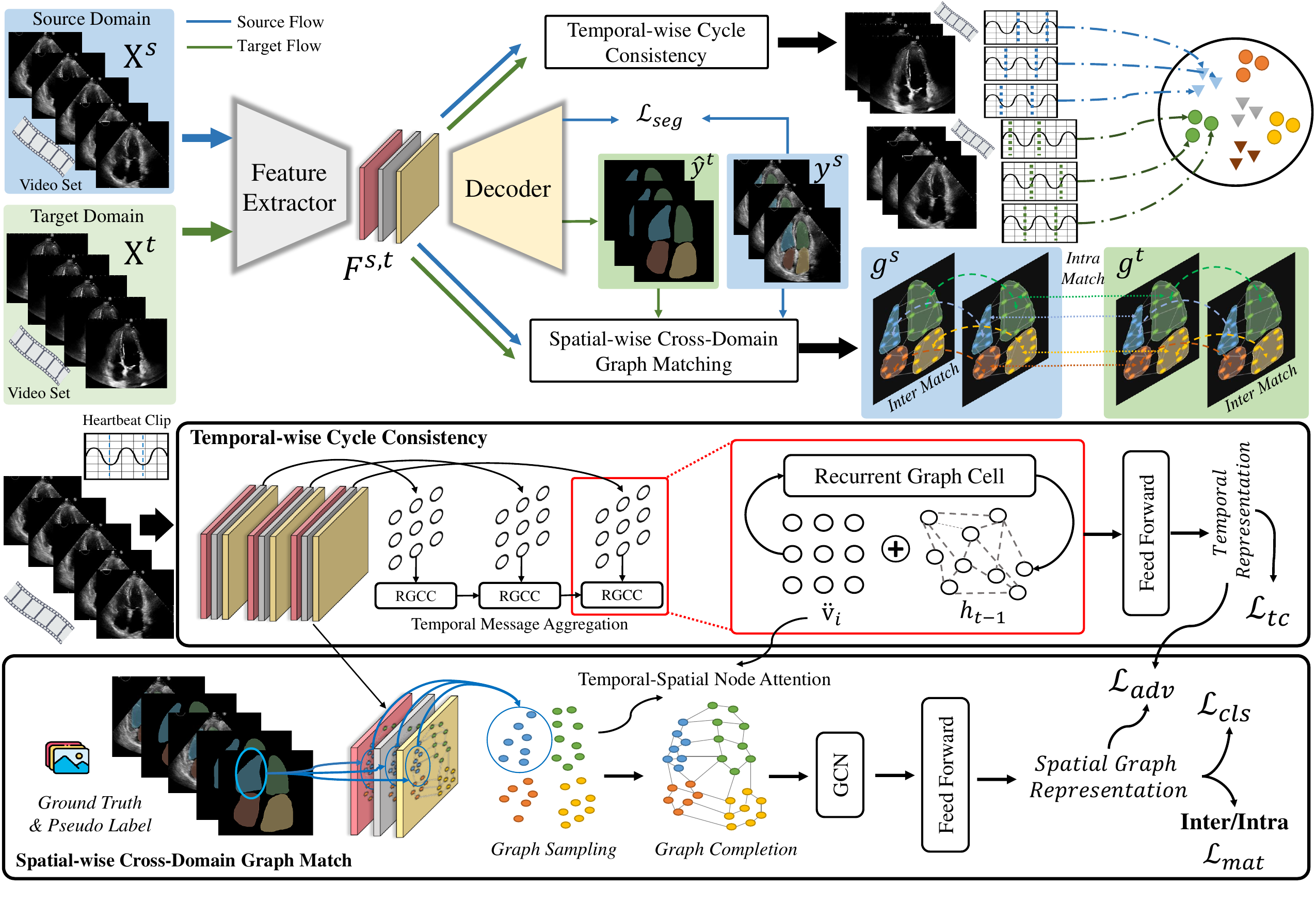}
  \caption{\textbf{The overview of our GraphEcho framework.} The source/target videos are fed into a shared backbone, generate feature maps, and produce the segmentation result from the decoder. Spatially, we extract feature nodes from each feature map based on their corresponding ground truth and pseudo label and, subsequently, construct complete semantic graphs suitable for intra-domain and inter-domain matching of cardiac structures. Temporally, we aggregate temporal messages by means of recurrent graph cells, thereby clustering heartbeat representations and enforcing temporal cycle consistency. Additionally, our method incorporates the temporal-spatial node attention module to establish a connection between the spatial and temporal domains. Finally, the trained feature extractor and decoder are used for inference and generating the final result.
  \label{Fig:overview_pipeline}}
  \end{center}
  \vspace{-15pt}
\end{figure*}

%
As shown in Figure~\ref{Fig:overview_pipeline}, our method consists of three main components.
First, a basic segmentation network is used to extract features and obtain prediction masks for both source and target domain data (See Section~\ref{sec:segmentation}).
Then, a \textbf{S}patial-wise \textbf{C}ross-domain \textbf{G}raph \textbf{M}atching (SCGM) module and a \textbf{T}emporal-wise \textbf{C}ycle \textbf{C}onsistency (TCC) module are designed to reduce the domain gap in both spatial and temporal wise for echocardiogram videos (See Section~\ref{sec:chamber-based Graph Matching} and (See Section~\ref{sec:Temporal-wise Cycle Consistency})).
%
\subsection{Basic Segmentation Network}~\label{sec:segmentation}
In our UDA echocardiogram segmentation, we denote the source and target domain data as $\{ \mathcal{X}^s, \mathcal{Y}^s \}$ and $\mathcal{X}^t$, respectively, where $\mathcal{X}^s$ is the video set in the source domain, and $\mathcal{Y}^s$ is its corresponding label set. Note that the videos set in the target domain $\mathcal{X}^t$ is without the label. 
For clarity, we sample a video frame with the label $\{ \mathbf{x}^s, \mathbf{y}^s \}$ from an example $\{ \mathbf{X}^s, \mathbf{Y}^s \}$ of the source domain data, where $\mathbf{X}^s \in \mathcal{X}^s$ is a video from $\mathcal{X}^s$ and $\mathbf{Y}^s \in \mathcal{Y}^s$ is its corresponding label.
%
%
Similarly, we can also sample a video frame from the target domain,~\ie, $\mathbf{x}^t$.

The basic segmentation network consists of a feature extractor and a decoder.
We first feed the $\mathbf{x}^s$ or $\mathbf{x}^t$ to the feature extractor to obtain $\mathbf{f}^s$ or $\mathbf{f}^s$, followed by a decoder that maps the features $\mathbf{f}^s$ or $\mathbf{f}^t$ to the corresponding prediction mask,~\ie, $\hat{\mathbf{y}}^s$ or $\hat{\mathbf{y}}^t$.
Then, we use the segmentation loss to supervise the model on the pixel classification task with the annotated source domain data as follows:
\begin{equation}
    \mathcal{L}_{seg} = \mathcal{L}_{bce}(\hat{\mathbf{y}}^s, {\mathbf{y}}^s) +  \mathcal{L}_{dice}(\hat{\mathbf{y}}^s, {\mathbf{y}}^s),
\end{equation}
where $\mathcal{L}_{bce}$ and $\mathcal{L}_{dice}$ are the binary cross-entropy loss and dice loss~\cite{milletari2016v}.

\subsection{Spatial-wise Cross-domain Graph Matching}~\label{sec:chamber-based Graph Matching}
%
%
%
In this section, we introduce the spatial-wise cross-domain graph matching  (SCGM), which aligns both class-wise representations and their relations across the source and target domains. 
To this end, we use the graph to model each echocardiogram frame, where the nodes represent the different chambers and the edges illustrate the relations between them.
Compared with the convolution neural network, the graph can better construct the relations among different classes explicitly~\cite{han2022vision}. 
%

In the following, we first illustrate how to convert the features of source and target domains~\emph{i.e.}, $\mathbf{f}^{s}$ and $\mathbf{f}^{s}$ to the corresponding graph representation, which is defined as $\mathbf{g}^{s}$ and $\mathbf{g}^{t}$ respectively. After that, we introduce the graph matching method to align the generated graph to reduce the domain gap.

\noindent\textbf{Graph construction.} The graph construction aims to convert visual features to graphs.
Since the construction process of the source domain and the target domain is the same, we take the source domain for illustration. 
Formally, given the feature $\mathbf{f}^s$ and its corresponding pseudo labels $\hat{\mathbf{y}}^s$ for a video frame (see definitions in Section~\ref{sec:segmentation}), we conduct graph sampling to sample the graph nodes from $\mathbf{f}^s$ based on $\hat{\mathbf{y}}^s$, as shown in Figure~\ref{Fig:overview_pipeline}. 
Specifically, we first use the pseudo labels $\hat{\mathbf{y}}^s$ to segment $\mathbf{f}^s$ into different chamber regions,~\ie, $\{ \mathbf{f}^s_i \}_{}$.  
Then, in each chamber region, we uniformly sample $m$ feature vectors fed into a projection layer to obtain the node embedding $\mathbf{v}^s$.
Based on $\mathbf{v}^s$, we define the edge connections $\mathbf{e}^{s}$, which is a learned matrix. Finally, the constructed semantic graph can be defined as $g^{s} = \{\mathbf{v}^s, \mathbf{e}^s \}$. In this same way, we can also obtain the semantic graph for the target domain,~\emph{i.e.}, $g^{t} = \{\mathbf{v}^t, \mathbf{e}^t \}$.

\noindent\textbf{Graph matching.} We leverage graph matching to perform the alignment of the source and target domain graph ${g}^s$ and ${g}^t$, thus reducing the domain gap.
Since graph matching is an optimization problem for ${g}^s$ and ${g}^t$, the relations between the two graphs are essential for the optimal solution~\cite{chen2020graph}.
Hence, we use the self-attention~\cite{vaswani2017attention,ding2022exploring} to capture the intra- and inter-domain relations between the source and target graph nodes,~\emph{i.e.}, $\mathbf{v}^{s}$ and $\mathbf{v}^{t}$, which can be formulated as $\bar{\mathbf{v}}^s, \bar{\mathbf{v}}^t = Attention(concat(\mathbf{v}^t,\mathbf{v}^s))$, where $concat$ indicates the concatenation.
To ensure the generated graph nodes are classified into the correct classes, we introduce the classification loss as follows:
\vspace{-5pt}
  \begin{equation}
    \mathcal{L}_{cls} = -\alpha  \mathbf{y} log(h(\bar{\mathbf{v}}^{s})) - \beta  \hat{\mathbf{y}} log(h(\bar{\mathbf{v}}^{t})),
    \label{eq:lcls}
    \vspace{-5pt}
  \end{equation}
where $h$ is the classifier head followed by a softmax, and $\alpha, \beta$ are the weights for two domains.
%
%

Then, graph matching can be implemented by maximising the similarity of graphs (including nodes and edges) belonging to the same class but from two different domains. 
Specifically, we first obtain the adjacency matrix $\mathbf{A}$ from $g^{s}$ and $g^{t}$ following to represent the relations of graph nodes. 
Then, the maximizing process can be transferred into optimizing the transport distance of $\mathbf{A}$. 
%
To this end, we use the Sinkhorn algorithm~\cite{cuturi2013sinkhorn} to obtain the transport cost matrix of graphs among chambers, defined as $\vec{\mathbf{A}} = Sinkhorn(\mathbf{A})$.
Then, our optimization target can be formulated as follows:
\vspace{-5pt}
\begin{equation}
  \begin{split}
      \mathcal{L}_{mat} =  \sum_{p,q} \; &\mathbb{I}(\mathbf{y}_p^s=\hat{\mathbf{y}}_q^t) \cdot min(\vec{\mathbf{A}}(p,q)) + \\
    &\mathbb{I}(\mathbf{y}_p^s\neq \hat{\mathbf{y}}_q^t) \cdot max(\vec{\mathbf{A}}(p,q)),
  \end{split}
  \vspace{-5pt}
  \label{eq:lmat}
\end{equation}
where $\mathbf{A}(p,q)$ is the $p$-th row and $q$-th column element on $\mathbf{A}$, $\mathbb{I}(\cdot)$ is the indicator function. 
Eq.~\ref{eq:lmat} aims to minimize the distance between samples of the same class across different domains while increasing the distance between samples of different classes across domains, thus eliminating the influence of domain shift.
Finally, $\mathcal{L}_{SCGM}=\mathcal{L}_{cls}+\mathcal{L}_{mat}$ is the overall loss of module SCGM.
%
%
\subsection{Temporal-wise Cycle Consistency} ~\label{sec:Temporal-wise Cycle Consistency}
In this section, we propose the \textbf{T}emporal \textbf{C}ycle \textbf{C}onsistency (TCC) module to enhance the temporal graphic representation learning across frames, by leveraging the temporal morphology of echocardiograms,~\emph{i.e.}, the discriminative heart cycle pattern across different patients.
The proposed TCC consists of three parts: a temporal graph node construction to
generate a sequence of temporal graph nodes for each video; a recursive graph convolutional cell to learn the global graph representations for each video; a temporal consistency loss to enhance the intra-video similarity and reduce the inter-video similarity. 
Note that TCC is applied to both source and target domains; we take the source domain for clarity.


\noindent\textbf{Temporal graph node construction.}
\label{sec:global_feature_nodes_sampling}
Given a video $\mathbf{X}^s$, we defines its features for frames as $\{ \mathbf{f}_i^s \}_{i=1}^N$, where $\mathbf{f}_i$ is the feature of the $i$-th frame and $N$ is the number of frames in $\mathbf{X}^s$.
Considering the computation cost, we use an average pooling layer to compress the size of $\{ \mathbf{f}_i^s \}_{i=1}^N$. For each compressed feature $\mathbf{f}_i^s$, we flatten it and treat its pixels as graphical nodes,~\ie, $\ddot{\mathbf{v}}_i^s$.
Then, the temporal graph nodes for the video $\mathbf{X}^s$ can be defined as $\{ \ddot{\mathbf{v}}_i^s \}_{i=1}^N$.
%


\begin{figure}[!t]
  \centering
  \includegraphics[width=0.99\linewidth]{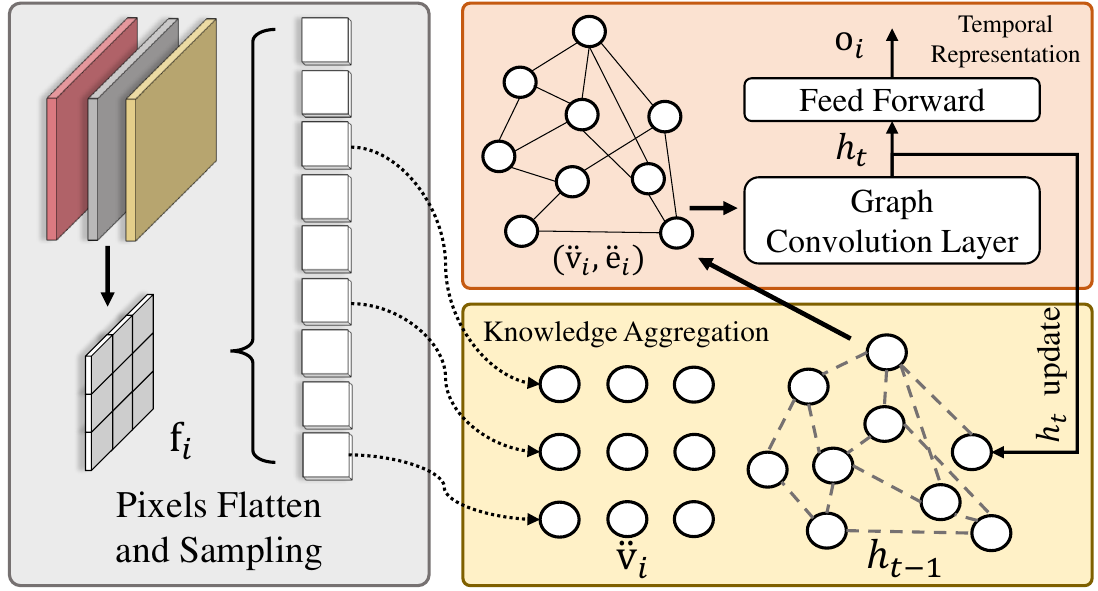}
  \caption{This figure illustrates the workflow of the Recursive Graph Convolutional Cell (RGCC), which receives the current input $\ddot{\mathbf{v}}^s_i$ and the hidden reference graph $h_{t-1}$. Complete the knowledge aggregation operation through the K-nearest neighbour algorithm that builds the edge $\ddot{\mathbf{e}}^s_i$ for $\ddot{\mathbf{v}}^s_i$. Finally, a graph convolution layer for nodes representation $\bar{x}_t$ calculation.}
  \label{fig:RGCN}
  \vspace{-7pt}
\end{figure}
\noindent\textbf{Recursive graph convolutional cell.}
Inspired by~\cite{yang2022recurring}, we propose the recursive graph convolutional cell to aggregate the semantics of the temporal graph nodes $\{ \ddot{\mathbf{v}}_i^s \}_{i=1}^N$ for obtaining the global temporal representation of each video.
For the $p$-th node $\ddot{\mathbf{v}}_i^s(p)$ at $\ddot{\mathbf{v}}_i^s$, we find its $K$ nearest neighbors $\mathcal{N}(p)$ on the hidden state $\mathbf{h}_t$ \footnote{$h_t$ is the learned parameters and the initial hidden state $h_0$ is all zero.}, where $\mathcal{N}(p) \in \mathbf{h}_t$.
Then the edge $\ddot{\mathbf{e}}^s_i({q,p})$ can be added directed from $\mathbf{h}_t(q)$ to $\ddot{\mathbf{v}}_i^s(p)$ for all $\mathbf{h}_t(q) \in \mathcal{N}(p)$.
After obtaining the edge $\ddot{\mathbf{e}}^s_i$ for $\ddot{\mathbf{v}}_i^s$, the message broadcast from the $i$-th graph to the $i+1$-th graph can be defined as follows:
\vspace{-3pt}
\begin{equation}
      \mathbf{h}_{t+1} = \sigma \mathbf{w}_{gcn}(\ddot{\mathbf{v}}^s_i, \ddot{\mathbf{e}}^s_i) + \mathbf{b}_{gcn},
\vspace{-3pt}
\end{equation}
where the $\sigma$ indicates the activation function, $ \mathbf{w}_{gcn}$ and $\mathbf{b}_{gcn}$ are the graph convolution weight and bias, respectively.
We conduct this message broadcast for $\{ \ddot{\mathbf{v}}^s_i, \ddot{\mathbf{e}}^s_i \}_{i=1}^N$, and obtain the final hidden state $\mathbf{h}_N$.
The global representation for video $\mathbf{X}^s$ is the $\mathbf{o}^s$, obtained by $\mathbf{o}^s = FFN (\mathbf{h}_N)$, where $FFN$ is a feed forward network.
Hence, the whole process of recursive graph convolutional cell can be formulated as $\mathbf{o}^s = RGCC(\mathbf{X}^s)$.
Similarly, we can obtain the temporal representation for the target video $\mathbf{X}^t$ by $\mathbf{o}^t = RGCC(\mathbf{X}^t)$.

\noindent\textbf{Temporal consistency loss.}
For better representation learning, we leverage temporal consistency loss to make features from the same video similar and features from different videos dissimilar.
In this paper, we use contrastive learning~\cite{han2021deep,ding2021support}, a mainstream method to pull close the positive pairs and push away negative ones, to achieve this goal.
We regard the two consequent clips $\mathbf{X}^s_{k}$ and $\mathbf{X}^s_{+}$ that randomly sampled from a video $\mathbf{X}^s$ as
positive pairs.
Then, we feed the positive clips to the recursive graph convolutional cell to obtain the global representations,~\ie, $\mathbf{o}^s_{k} = RGCC(\mathbf{X}^s_{k})$ and $\mathbf{o}^s_{+} = RGCC(\mathbf{X}^s_{+})$.
For negative pairs, we maintain a memory bank $\mathcal{B}$ consisting of representations of clips sampled from different videos.
Then, the temporal consistency loss for the source domain is defined as follows:
\begin{equation}
     \mathcal{L}_{tc}^s = -\sum_{ \{ \mathbf{o}^s_{k}, \mathbf{o}^s_{+}\} \in \mathcal{P}^s} log\frac{\text{exp}(\mathbf{o}^s_{k} \cdot \mathbf{o}^s_{+})}{\sum_{ \mathbf{o}^s_{-} \in \mathcal{B}} \text{exp}(\mathbf{o}^s_{k} \cdot \mathbf{o}^s_{-})},
  \label{eq:ltc}
  \end{equation}
where $\mathcal{P}^s$ is the set of positive pairs.
We here use the dot product to measure the similarity and use InfoNCE~\cite{oord2018representation} as the specific contrastive learning objective.
Similarly, we can define the temporal consistency loss for the target domain as $\mathcal{L}_{tc}^t$, and the total temporal consistency loss is $\mathcal{L}_{tc} = \mathcal{L}_{tc}^s + \mathcal{L}_{tc}^t$.

Since $\mathcal{L}_{tc}$ is applied to two domains independently, a gap between source and target domains  still exists for the learned global representation,~\ie, $\mathbf{o}^s$ or $\mathbf{o}^t$.
Hence, we leverage the adversarial methods~\cite{ganin2016domain} to eliminate the gap between $\mathbf{o}^s$ and $\mathbf{o}^t$, which can be formulated as $\mathcal{L}_{adv}$.
The overall loss of temporal consistency is $\mathcal{L}_{TCC} = \mathcal{L}_{tc}+\mathcal{L}_{adv}$, where $\mathcal{L}_{adv}$ is the global domain-adversarial loss in our TCC module.
To summarize, the final loss of GraphEcho is $\mathcal{L}_{All} = \mathcal{L}_{SCGM} + \mathcal{L}_{TCC} + \mathcal{L}_{seg}$, and the network is trained in end-to-end.
\section{Experiments}
\subsection{Datasets.}
We evaluate our method on three datasets, including our collected dataset (CardiacUDA) and two public datasets (CAMUS~\cite{leclerc2019deep} and Echonet Dynamic~\cite{ouyang2020video}).
Table~\ref{tab:dataset_comp} shows the dataset details.

\noindent\textbf{CardiacUDA.} 
We collect CardiacUDA from our two collaborating hospitals: site G and site R. In order to guarantee all echocardiogram videos are standards-compliant, all cases of CardiacUDA are collected, annotated and approved by 5-6 experienced physicians. For ethical issues, we have required approval from medical institutions.

Each patient underwent four views during scanning, which included parasternal left ventricle long axis (LVLA), pulmonary artery long axis (PALA), left ventricular short-axis (LVSA), and apical four-chamber heart (A4C), resulting in four videos per patient. The resolution of each video was either 800x600 or 1024x768, depending on the scanner used (Philips or HITACHI). A total of 516 and 476 videos were collected from Site G and Site R, respectively, from approximately 100 different patients. Each video consists of over 100 frames, covering at least one heartbeat cycle.

We have provided pixel-level annotations for each view, including masks for the left ventricle (LV) and right ventricle (RV) in the LVLA view, masks for the pulmonary artery (PA) in the PALA view, masks for the LV and RV in the LVSA view, and masks for the LV, RV, left atrium (LA), and right atrium (RA) in the A4C view. 
The videos in both Site R and Site G were divided into a ratio of 8:1:1 for training, validation, and testing, respectively. To lower annotation costs in the training set, only five frames per video are provided with pixel-level annotation masks.
To better measure the model performance, we provide pixel-level annotations for every frame in each video in the validation and testing sets. 



%



\noindent\textbf{CAMUS~\cite{leclerc2019deep}} consists of $500$ echocardiogram videos with pixel-level annotations for the left ventricle, myocardium, and left atrium.
To save the annotation cost, only 2 frames (end diastole and end systole) are labelled in each video. We randomly split the dataset into $8:1:1$ for training, validation, and testing. 

\noindent\textbf{Echonet Dynamic~\cite{ouyang2020video}} is the largest echocardiogram video dataset, including 10,030 videos with human expert annotations. 
Similarly, we split videos into $8:1:1$ for training, validation, and testing, respectively.

\subsection{Implementation Details}
\noindent\textbf{Training.} 
All methods are built on the ``DeepLabv3''~\cite{chen2017rethinking} backbone for fair comparison. We trained the model using the stochastic gradient descent (SGD) optimizer with a weight decay of $0.0001$ and a momentum of $0.9$. The model was trained for a total of 400 epochs with an initial learning rate of $0.02$, and the learning rate was decreased by a factor of 0.1 every 100 epochs. The batch size was set to 4.
For spatial data augmentation, each frame was resized to $384\times384$ and then randomly cropped to $256\times256$. The frames were also randomly flipped vertically and horizontally. As for temporal data augmentation, we randomly selected 40 frames from an echocardiogram video and sampled 10 frames as input equidistantly.
We followed the same training and data augmentation approach for the CAMUS and Echonet dynamic datasets as we did for our dataset.


%
%
\begin{table*}[!t]
  \centering
  \caption{Results on {CardiacUDA} dataset.
   ``\textit{Without DA}'': evaluating the model trained on the \textit{source domain} directly on the target domain. 
   ``\textit{Upper Bound}'': evaluating the model trained on the \textit{target domain} with labels directly on the same domain. ``Avg.'' refers to the averaged Dice score over four views, including LVLA, PALA, LVSA, and A4C. All results are reported in Dice score (\%)
  }
    \begin{tabular}{ccccccccccc}
\hlineB{3}
\multicolumn{1}{c|}{\multirow{2}{*}{Method}} & \multicolumn{5}{c|}{site G (\textit{source}) $\longrightarrow$ site R (\textit{target})} & \multicolumn{5}{c}{site R \textit{(source)} $\longrightarrow$ site G \textit{(target)}} \\ \cline{2-11} 
\multicolumn{1}{c|}{}                        & LVLA  & PALA  & LVSA  & A4C & \multicolumn{1}{c|}{Avg.} & LVLA       & PALA       & LVSA       &   A4C  & Avg.  \\ \hline
\multicolumn{11}{c}{Semi-Supervised Segmentation Methods} \\ \hline
\multicolumn{1}{c|}
{CPS~\cite{chen2021semi}}          & 63\tiny{.2$\pm$0.9}   & 60\tiny{.5$\pm$1.1}   & 57\tiny{.0$\pm$0.8}  & 64\tiny{.2$\pm$1.0} & \multicolumn{1}{c|}{61\tiny{.2$\pm$0.7}}  & 64\tiny{.9$\pm$1.4}        & 63\tiny{.3$\pm$1.1}        & 59\tiny{.8$\pm$0.6}        & 61\tiny{.0$\pm$1.9}    & 62\tiny{.2$\pm$1.4}    \\ 
\multicolumn{1}{c|}{PC$^2$Seg~\cite{zhong2021pixel}}   & 64\tiny{.1$\pm$0.8}   & 58\tiny{.2$\pm$0.6}   & 70\tiny{.2$\pm$0.9}  & 63\tiny{.9$\pm$1.2} & \multicolumn{1}{c|}{64\tiny{.1$\pm$1.1}}  & 63\tiny{.7$\pm$1.0}        & 64\tiny{.4$\pm$1.0}        & 65\tiny{.2$\pm$1.3}        & 67\tiny{.0$\pm$0.9}    & 65\tiny{.1$\pm$0.8}    \\
\multicolumn{1}{c|}{U$^2$PL~\cite{wang2022semi}}  &  66\tiny{.2$\pm$1.2} &  62\tiny{.9$\pm$1.5}  &  69\tiny{.1$\pm$0.9} & 64\tiny{.4$\pm$1.4} & \multicolumn{1}{c|}{65\tiny{.6$\pm$1.6}}  &  62\tiny{.1$\pm$1.2} &    61\tiny{.5$\pm$1.1}   &  61\tiny{.9$\pm$1.0}    &  65\tiny{.0$\pm$0.5}  &  62\tiny{.7$\pm$0.9}     \\ \hline 
\multicolumn{11}{c}{Unsupervised Domain Adaptation Methods} \\ \hline
\multicolumn{1}{c|}{\textit{Without DA}}          & 53\tiny{.5$\pm$0.4}   & 42\tiny{.8$\pm$0.7}   & 47\tiny{.6$\pm$0.6}  & 50\tiny{.1$\pm$1.0}  & \multicolumn{1}{c|}{48\tiny{.5$\pm$0.7}}  & 55\tiny{.2$\pm$0.5}        & 47\tiny{.4$\pm$0.6}        & 49\tiny{.1$\pm$0.3}        & 52\tiny{.8$\pm$0.9}     & 51\tiny{.1$\pm$0.5}  \\
\multicolumn{1}{c|}{CCM~\cite{li2020content}}     & 13\tiny{.5$\pm$2.6}   & 10\tiny{.3$\pm$3.8}   & 13\tiny{.0$\pm3.2$}  & 20\tiny{.7$\pm$2.9}  & \multicolumn{1}{c|}{22\tiny{.3$\pm$}3.5}  & 17\tiny{.8$\pm$4.0}        & 14\tiny{.2$\pm$3.1}        & 25\tiny{.7$\pm$2.9}        & 16\tiny{.6$\pm$3.2}    & 18\tiny{.6$\pm$3.0}   \\
\multicolumn{1}{c|}{Caco~\cite{huang2022category}}  &   40\tiny{.7$\pm$2.4} &   39\tiny{.9$\pm$2.6} &  28\tiny{.4$\pm$2.1}  &  34\tiny{.1$\pm$1.9} & \multicolumn{1}{c|}{35\tiny{.8$\pm$}2.3}  &  36\tiny{.2$\pm$3.0} &      35\tiny{.9$\pm$2.6} &   27\tiny{.0$\pm$1.8}   &   38\tiny{.3$\pm$2.5}  & 32\tiny{.7$\pm$2.8} \\
\multicolumn{1}{c|}{RIPU~\cite{xie2022towards}}   & 36\tiny{.0$\pm$2.3}   & 34\tiny{.9$\pm$2.8}   & 25\tiny{.8$\pm$3.0}  & 31\tiny{.1$\pm$1.9} & \multicolumn{1}{c|}{31\tiny{.9$\pm$2.2}}  & 27\tiny{.4$\pm$2.1}        & 36\tiny{.2$\pm$1.7}       & 32\tiny{.5$\pm$1.9}        & 34\tiny{.9$\pm$2.6}    & 32\tiny{.8$\pm$2.4}    \\ 
\multicolumn{1}{c|}{CPSL~\cite{li2022class}}      & 35\tiny{.4$\pm$1.4}   & 45\tiny{.1$\pm$1.6}   & 39\tiny{.7$\pm$1.5}  & 51\tiny{.2$\pm$1.3}  & \multicolumn{1}{c|}{42\tiny{.6$\pm$2.0}}  & 44\tiny{.2$\pm$1.6}   & 53\tiny{.0$\pm$2.3}        & 39\tiny{.5$\pm$1.5}   &   42\tiny{.6$\pm$2.0}  & 44\tiny{.8$\pm$1.9} \\  
\multicolumn{1}{c|}
{PLCA~\cite{kang2020pixel}}                       & 58\tiny{.2$\pm$1.7}   & 21\tiny{.0$\pm$4.9}   & 40\tiny{.2$\pm$3.6}  & 60\tiny{.3$\pm$2.3}  & \multicolumn{1}{c|}{44\tiny{.9$\pm$3.1}}  & 60\tiny{.1$\pm$2.9}   & 38\tiny{.8$\pm$3.0}        & 42\tiny{.9$\pm$2.7}        & 59\tiny{.4$\pm$2.6}   & 50\tiny{.3$\pm$2.9}    \\
\multicolumn{1}{c|}{PixMatch~\cite{melas2021pixmatch}}  & 60\tiny{.8$\pm$1.8}   & 52\tiny{.7$\pm$1.6}   & 56\tiny{.0$\pm$1.8}  & 66\tiny{.5$\pm$1.5}  & \multicolumn{1}{c|}{59\tiny{.0$\pm$}1.7}  & 62\tiny{.9$\pm$2.0}       & 49\tiny{.0$\pm$3.2}        & 63\tiny{.2$\pm$2.1}        & 69\tiny{.9$\pm$1.8}    & 61\tiny{.3$\pm$1.8}   \\
\multicolumn{1}{c|}{FDA~\cite{yang2020fda}}       & 67\tiny{.3$\pm$2.0}   & 65\tiny{.5$\pm$1.5}   & 54\tiny{.8$\pm$2.3}  & 64\tiny{.3$\pm$1.9} & \multicolumn{1}{c|}{63\tiny{.0$\pm$1.5}}  & 65\tiny{.8$\pm$1.7}        & 63\tiny{.2$\pm$1.8}        & 61\tiny{.9$\pm$2.1}        & 64\tiny{.5$\pm$1.5}     & 63\tiny{.9$\pm$1.8}  \\
\multicolumn{1}{c|}{FDA-MBT~\cite{yang2020fda}}   & 64\tiny{.4$\pm$0.9}   & 65\tiny{.1$\pm$0.8}   & 61\tiny{.7$\pm$1.1}  & 70\tiny{.1$\pm$1.3}  & \multicolumn{1}{c|}{65\tiny{.3$\pm$1.2}}  & 66\tiny{.3$\pm$0.9}        & 64\tiny{.9$\pm$1.4}   & 67\tiny{.2$\pm$0.6}        & 71\tiny{.3$\pm$0.9}    & 67\tiny{.4$\pm$0.8}   \\
\multicolumn{1}{c|}{FADA~\cite{wang2020classes}}  & 70\tiny{.1$\pm$1.2}   & 68\tiny{.3$\pm$1.4}   & 76\tiny{.1$\pm$0.7}  & 72\tiny{.4$\pm$0.6} & \multicolumn{1}{c|}{71\tiny{.7$\pm$0.8}}  & 69\tiny{.9$\pm$0.9}        & 67\tiny{.7$\pm$1.0}        & 74\tiny{.5$\pm$1.4}        & 70\tiny{.0$\pm$0.5}    & 70\tiny{.5$\pm$1.1}    \\ 
\multicolumn{1}{c|}{Ours}        & \textbf{73\tiny{.9$\pm$1.2}}   & \textbf{75\tiny{.5$\pm$1.3}}   & \textbf{76\tiny{.8$\pm$0.4}}  & \textbf{76\tiny{.3$\pm$0.7}}  & \multicolumn{1}{c|}{\textbf{75\tiny{.6$\pm$0.9}}}  & \textbf{73\tiny{.3$\pm$1.0}}        & \textbf{74\tiny{.9$\pm$1.2}}        & \textbf{80\tiny{.2$\pm$0.3}}        & \textbf{78\tiny{.2$\pm$0.5}}   & \textbf{76\tiny{.7$\pm$0.5}}     \\
\hline 
\multicolumn{1}{c|}{\textcolor{gray}{\textit{Upper Bound}}}  & \textcolor{gray}{79\tiny{.1$\pm$0.4}}   & \textcolor{gray}{82\tiny{.4$\pm$0.6}}   & \textcolor{gray}{82\tiny{.1$\pm$1.0}}  & \textcolor{gray}{81\tiny{.4$\pm$1.2}}  & \multicolumn{1}{c|}{\textcolor{gray}{81\tiny{.3$\pm$0.5}}}  & \textcolor{gray}{80\tiny{.5$\pm$1.4}}        & \textcolor{gray}{79\tiny{.2$\pm$0.8}}        & \textcolor{gray}{83\tiny{.3$\pm$1.6}}        & \textcolor{gray}{83\tiny{.9$\pm$0.2}}  & \textcolor{gray}{81\tiny{.7$\pm$0.8}}       \\ \hlineB{3}
\end{tabular}
  \label{tab:Result_GY_RMYY}
\end{table*}
  \begin{table*}[!t]
  \centering
  \caption{Results on CAMUS, Echonet dynamic and CardiacUDA datasets. As only LV segmentation labels are provided in these three datasets, we report the results on the dice score of LV segmentation.
  ``EDV''  and ``ESV'' refers to the Dice score of LV segmentation results at end-systole and end-diastole frames, respectively. All results are reported in Dice score (\%)}
  \begin{adjustbox}{width=0.999\textwidth}
    \begin{tabular}{c|cc|cc|cc|cc|c|c}
\hlineB{3}
\multirow{2}{*}{Method} & \multicolumn{2}{c|}{CAMUS→Echo}   & \multicolumn{2}{c|}{Echo→CAMUS}   & \multicolumn{2}{c|}{Ours→Echo}   & \multicolumn{2}{c|}{Ours→CAMUS}  & \multicolumn{1}{c|}{Echo→Ours}  & \multicolumn{1}{c}{\footnotesize{CAMUS→Ours}}\\ \cline{2-11} 
                    & \multicolumn{1}{c|}{EDV} & ESV   & \multicolumn{1}{c|}{EDV} & ESV   & \multicolumn{1}{c|}{EDV} & ESV   & \multicolumn{1}{c|}{EDV} & ESV   & \multicolumn{1}{c|}{Avg.}   & \multicolumn{1}{c}{Avg.}  \\ \hline
\textit{\footnotesize{Without DA.}}      &   69\tiny{.2$\pm$1.4}   &   66\tiny{.2$\pm$2.2}   &   64\tiny{.3$\pm$0.9}    &   59\tiny{.6$\pm$1.7}   &   34\tiny{.1$\pm$1.4}   &   33\tiny{.8$\pm$2.2}   &   31\tiny{.0$\pm$0.9}    &   32\tiny{.4$\pm$1.7}   &   22\tiny{.9$\pm$1.6}    &   19\tiny{.2$\pm$1.4}\\
\footnotesize{U$^2$PL~\cite{wang2022semi}}            &   63\tiny{.2$\pm$0.9}   &   67\tiny{.8$\pm$0.9}   &   57\tiny{.2$\pm$1.1}    &   60\tiny{.1$\pm$1.2}   &   49\tiny{.5$\pm$0.9}   &   51\tiny{.3$\pm$0.7}   &   43\tiny{.1$\pm$0.9}    &   46\tiny{.7$\pm$1.2}   &   36\tiny{.5$\pm$1.0}    &   34\tiny{.3$\pm$0.8}\\
\footnotesize{Caco~\cite{huang2022category}}               &   55\tiny{.9$\pm$1.5}   &   56\tiny{.0$\pm$1.3}   &   47\tiny{.3$\pm$1.3}    &   49\tiny{.0$\pm$1.2}   &   38\tiny{.6$\pm$3.4}   &   40\tiny{.8$\pm$2.6}   &   46\tiny{.1$\pm$1.9}    &   45\tiny{.5$\pm$2.2}   &   29\tiny{.6$\pm$1.8}    &   26\tiny{.8$\pm$2.3}\\ 
\footnotesize{RIPU~\cite{xie2022towards}}               &   64\tiny{.3$\pm$1.2}   &   67\tiny{.7$\pm$1.0}   &   70\tiny{.2$\pm$0.6}    &   68\tiny{.2$\pm$0.9}   &   46\tiny{.9$\pm$0.7}   &   47\tiny{.4$\pm$0.7}   &   56\tiny{.0$\pm$1.2}    &   51\tiny{.5$\pm$1.4}   &   36\tiny{.2$\pm$1.7}    &   31\tiny{.7$\pm$2.0}\\ 
\footnotesize{PLCA~\cite{kang2020pixel}}              &   71\tiny{.1$\pm$0.4}   &   69\tiny{.3$\pm$0.5}   &   72\tiny{.9$\pm$0.8}    &   68\tiny{.3$\pm$0.7}   &   51\tiny{.9$\pm$0.9}   &   49\tiny{.7$\pm$0.9}   &   52\tiny{.4$\pm$1.4}    &   52\tiny{.1$\pm$1.2}   &   35\tiny{.3$\pm$1.1}    &   36\tiny{.1$\pm$0.8}\\ 
\footnotesize{FDA~\cite{yang2020fda}}               &   78\tiny{.8$\pm$1.1}   &   75\tiny{.4$\pm$1.0}   &   76\tiny{.2$\pm$0.4}    &   74\tiny{.1$\pm$0.5}   &   55\tiny{.6$\pm$0.4}   &   54\tiny{.0$\pm$0.6}   &   56\tiny{.8$\pm$0.5}    &   56\tiny{.1$\pm$0.3}   &   38\tiny{.4$\pm$1.5}    &   37\tiny{.9$\pm$1.2}\\ 
\footnotesize{FADA~\cite{wang2020classes}}               &   77\tiny{.5$\pm$0.8}   &   76\tiny{.5$\pm$0.5}   &   78\tiny{.6$\pm$0.8}    &   76\tiny{.6$\pm$0.6}   &   54\tiny{.1$\pm$0.6}   &   52\tiny{.0$\pm$1.0}   &   57\tiny{.4$\pm$1.0}    &   55\tiny{.2$\pm$0.8}   &   41\tiny{.7$\pm$1.1}    &   39\tiny{.0$\pm$0.9}\\ 
\textbf{Ours}                   &   \textbf{83\tiny{.4$\pm$0.7}}   &   \textbf{81\tiny{.8$\pm$0.9}}   &   \textbf{87\tiny{.6$\pm$0.4}}    &   \textbf{82\tiny{.4$\pm$1.0}}   &   \textbf{61\tiny{.2$\pm$0.5}}   &   \textbf{61\tiny{.8$\pm$0.7}}    &   \textbf{66\tiny{.3$\pm$0.3}}   &   \textbf{64\tiny{.9$\pm$0.4}}   &   \textbf{46\tiny{.2$\pm$1.3}}    &   \textbf{44\tiny{.0$\pm$1.6}}\\ 
\textcolor{gray}{\textit{\scriptsize{Upper Bound}}}   & \textcolor{gray}{93\tiny{.4$\pm$0.6}}   & \textcolor{gray}{90\tiny{.5$\pm$1.3}}   & \textcolor{gray}{89\tiny{.3$\pm$1.1}}   & \textcolor{gray}{87\tiny{.9$\pm$0.8}}   & \textcolor{gray}{93\tiny{.4$\pm$0.6}}   & \textcolor{gray}{90\tiny{.5$\pm$1.3}}   & \textcolor{gray}{89\tiny{.3$\pm$1.1}}   & \textcolor{gray}{87\tiny{.9$\pm$0.8}}   & \textcolor{gray}{81\tiny{.3$\pm$0.9}}   & \textcolor{gray}{81\tiny{.3$\pm$0.6}}\\ \hlineB{3}
\end{tabular}
  \end{adjustbox}
  \label{tab:Result_Echo_CAMUS}
  \vspace{-10pt}
  \end{table*}

\noindent\textbf{Validation and Testing.}
We chose the model with the highest performance on the validation set and reported its results on the testing set. During the inference stage, we only used center cropping as the preprocessing.

\subsection{Comparison with the State-of-the-Art Methods}
%

\noindent\textbf{Results on CardiacUDA.}
We compare our method with existing state-of-the-art UDA methods~\cite{li2020content,li2022class,huang2022category,kang2020pixel,melas2021pixmatch,yang2020fda,yang2020fda,wang2020classes,xie2022towards} in the computer vision domain. Furthermore, considering the similar visual appearances of different domains, we also compare our method with several state-of-the-art semi-supervised segmentation methods~\cite{chen2021semi,zhong2021pixel,wang2022semi}, where images in the source domain are considered as labelled images, and images from target domain are treated as unlabelled images.



\begin{figure*}[!t]
  \begin{center}
  \includegraphics[width=0.999\linewidth]{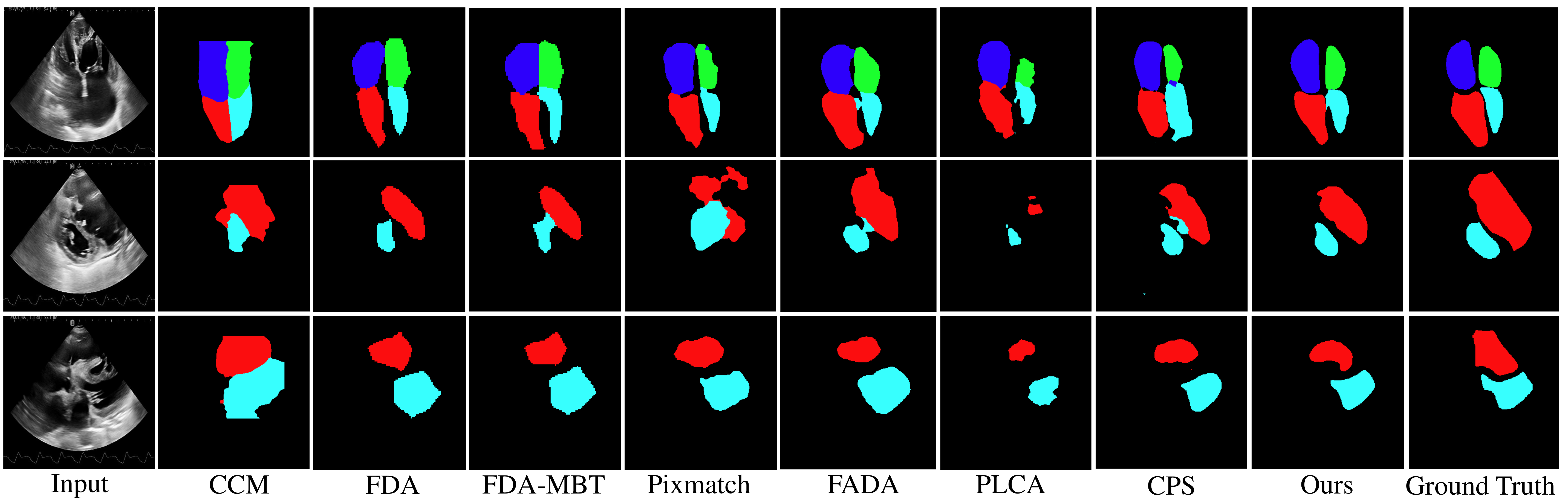}
  \caption{We visualize three video frames to show the segmentation results. Red, green, blue, and cyan indicate refer to the segmentation regions for the right Atrium (RA), left ventricle (LV), right ventricle (RV), and left atrium (LV), respectively.}
  \vspace{-20pt}
  \label{Fig:visulization}
  \end{center}
\end{figure*}

The performance is evaluated on two settings, as shown in Table~\ref{tab:Result_GY_RMYY}. 
Our method demonstrated superior performance compared to the best-performing method~\cite{wang2020classes}, achieving a 3.9\% and 6.2\% improvement on averaged Dice under two settings, respectively.
%
Notably, our method can surpass the best semi-supervised segmentation methods~\cite{wang2022semi,zhong2021pixel} by 10\% and 11.6\% on averaged Dice under two settings, respectively. This comparison further highlights the significant domain gaps between site G and site R, demonstrating the effectiveness of our developed UDA method. 
Figure~\ref{Fig:visulization} shows the visualization of the segmentation results, where our method outperformed the other methods.

\noindent\textbf{Results on our CardiacUDA, CAMUS, and Echonet.} 
Table~\ref{tab:Result_Echo_CAMUS} shows the results of our UDA methods under six settings with three datasets.
$a\rightarrow b$ indicates that $a$ is the source domain and $b$ is the target domain. We can see that our method can achieve excellent performance under six settings. Notably, as shown in Echo $\rightarrow$ CAMUS, our method can achieve 87.6\% and 82.4\% on Dice for EDV and ESV, respectively, which are very close to the upper bound of this setting. We also compare our method with state-of-the-art methods on different settings in Table~\ref{tab:Result_Echo_CAMUS}, which shows our method outperforms all other methods with significant improvement.

\begin{table*}[!t]
    \hspace{3pt}
    \begin{minipage}[t]{.32\linewidth}
      \caption{Effectiveness of SCGM and TCC. }
      \vspace{3pt}
      \setlength{\tabcolsep}{5.9pt}
      \resizebox{0.99\textwidth}{!}{
      \begin{tabular}{c|c|c|c|c}
        \hlineB{3}
            \multirow{2}*{} &  \multirow{2}*{SCGM} &  \multirow{2}*{TCC}  &\multicolumn{2}{c}{Dice Scores (\%)} \\ 
            \cline{4-5}
             & &  & G→R       & R→G \\
    	\hline
    	Base &  \tikzxmark  & \tikzxmark   & 48\tiny{.5$\pm$0.7} & 51\tiny{.1$\pm$0.5}\\
            Base + SCGM & \tikzcmark & \tikzxmark  & 74\tiny{.3$\pm$1.0} & 71\tiny{.3$\pm$0.7}  \\
            Base + TCC &  \tikzxmark  &  \tikzcmark & 55\tiny{.3$\pm$0.8} & 53\tiny{.0$\pm$1.2} \\
            Ours & \tikzcmark & \tikzcmark & \textbf{75\tiny{.6$\pm$0.9}} & \textbf{76\tiny{.7$\pm$0.5}}\\
        \hlineB{3}
    \end{tabular}
    }
    \label{tab:effectofmodule}
    \end{minipage}%
    \hspace{5pt}
    \begin{minipage}[t]{.32\linewidth}
      \centering
        \caption{Effect of $\mathcal{L}_{cls}$ (Eq.~\ref{eq:lcls}) and $\mathcal{L}_{mat}$ (Eq.~\ref{eq:lmat}) in SCGM.}
        \vspace{3pt}
        \setlength{\tabcolsep}{15pt}
        \resizebox{0.99\textwidth}{!}{
        \begin{tabular}{c|c|c|c}
        \hlineB{3.5}
	\multirow{2}*{$\mathcal{L}_{cls}$} &  \multirow{2}*{$\mathcal{L}_{mat}$}  &\multicolumn{2}{c}{Averaged Dice Score (\%)} \\ 
        \cline{3-4} &  & G→R       & R→G \\ \hline
        \tikzxmark & \tikzxmark   & 48\tiny{.5$\pm$0.7} & 51\tiny{.1$\pm$0.5} \\
	\tikzcmark & \tikzxmark   & 51\tiny{.9$\pm$1.2} & 53\tiny{.7$\pm$1.3} \\
        \tikzxmark & \tikzcmark   & 53\tiny{.4$\pm$1.6} & 54\tiny{.0$\pm$1.5} \\
        \tikzcmark & \tikzcmark   & 74\tiny{.3$\pm$1.0} & 71\tiny{.3$\pm$0.7}\\
        \hlineB{3.5}
	\end{tabular}
    }
    \label{tab:effectofscgm}
    \end{minipage}%
    \hspace{5pt}
    \begin{minipage}[t]{.32\linewidth}
      \centering
      \caption{Effect of $\mathcal{L}_{tc}$ (Eq.~\ref{eq:ltc}) and $\mathcal{L}_{adv}$ in TCC.}
      \vspace{2pt}
      \vspace{3pt}
      \setlength{\tabcolsep}{16pt}
      \resizebox{0.99\textwidth}{!}{
        \begin{tabular}{c|c|c|c}
        \hlineB{3}
	\multirow{2}*{$\mathcal{L}_{tc}$} &  \multirow{2}*{$\mathcal{L}_{adv}$}  &\multicolumn{2}{c}{Averaged Dice Score (\%)} \\ \cline{3-4}
        &  & G→R       & R→G \\ \hline
        \tikzxmark & \tikzxmark  & 74\tiny{.3$\pm$0.7} & 71\tiny{.3$\pm$0.5} \\
	\tikzcmark & \tikzxmark  & 74\tiny{.4$\pm$1.1} & 75\tiny{.6$\pm$0.9} \\
        \tikzxmark & \tikzcmark  & 74\tiny{.1$\pm$0.9} & 73\tiny{.5$\pm$1.0} \\
        \tikzcmark & \tikzcmark  & \textbf{75\tiny{.6$\pm$0.9}} & \textbf{76\tiny{.7$\pm$0.5}} \\
        \hlineB{3}
        \end{tabular}
        }
    \label{tab:tcc}
    \end{minipage} 
    \vspace{-10pt}
\end{table*}
\subsection{Ablation Study}
\noindent\textbf{Effectiveness of SCGM and TCC.}
Table~\ref{tab:effectofmodule} shows the effectiveness of our proposed SCGM and TCC.
``Base" indicates the basic segmentation network.
The results show that adopting SCGM can largely improve the base model from 48.5\% to 74.3\% under setting G $\rightarrow$ R. However, only applying TCC shows limited improvements over the base model. This is mainly because the TCC is designed to jointly train unlabelled data and construct better graphical representation in a temporal manner, which does not include any operation that focuses on narrowing the domain discrepancy, leading to limited adaptation results.
The combination of SCGM and TCC can achieve the best performance.

\noindent\textbf{Ablation study of SCGM.} Since there are two loss functions,~\ie, $\mathcal{L}_{cls}$ (Eq.~\ref{eq:lcls}) and $\mathcal{L}_{mat}$ (Eq.~\ref{eq:lmat}) in SCGM, we ablate their effects in Table~\ref{tab:effectofscgm}.
The results illustrate that using $\mathcal{L}_{cls}$ and $\mathcal{L}_{mat}$ alone can only achieve limited improvements.
This is because only using $\mathcal{L}_{cls}$ can not align the representations from different domains well while only using $\mathcal{L}_{mat}$ may perform the erroneous alignment,~\eg, align the features of LV to those of RV.
By combining two losses, we can conduct the correct class-wise alignment and achieve significant improvement.
  \begin{table}[t!]
    \centering
    \caption{Analysis of different attentions. ``None'' denotes that no attention module is applied in our framework, while the ``Inter'', ``Intra'', and ``Inter-Intra'' refers to cross-domain, internal domain, and dual (cross$+$internal) attention, respectively.}
    \begin{adjustbox}{width=0.482\textwidth}
        \begin{tabular}{c|ccccc}
\hlineB{3}
\multirow{2}{*}{Method}     & \multicolumn{5}{c}{Site G $\longrightarrow$ Site R}    \\ 
                            \cline{2-6} 
                            & LVLA      & PALA      & LVSA      & A4C     & Avg.     \\ 
                            \hline
None                        & 68\tiny{.1$\pm$2.2}      & 69\tiny{.8$\pm$0.9}      & 71\tiny{.4$\pm$1.3}      & 73\tiny{.3$\pm$0.4}    & 70\tiny{.7$\pm$1.2}     \\
Inter                       & 70\tiny{.5$\pm$1.5}      & 72\tiny{.6$\pm$1.3}      & 72\tiny{.0$\pm$0.8}      & 73\tiny{.5$\pm$0.4}    & 72\tiny{.2$\pm$1.0}     \\
Intra                       & 72\tiny{.1$\pm$0.5}      & 71\tiny{.5$\pm$0.8}      & 75\tiny{.2$\pm$1.7}      & 74\tiny{.6$\pm$0.7}    & 73\tiny{.4$\pm$0.9}     \\
{Inter-Intra }              & \textbf{73\tiny{.9$\pm$1.2}}   & \textbf{75\tiny{.5$\pm$1.3}}   & \textbf{76\tiny{.8$\pm$0.4}}  & \textbf{76\tiny{.3$\pm$0.7}}  & \textbf{75\tiny{.6$\pm$0.9}}     \\ 
                            \hlineB{3}
\end{tabular}
    \end{adjustbox}
    \label{tab:abla_attn}
  \end{table}
\begin{figure}[!t]
  \begin{center}
  \includegraphics[width=0.999\linewidth]{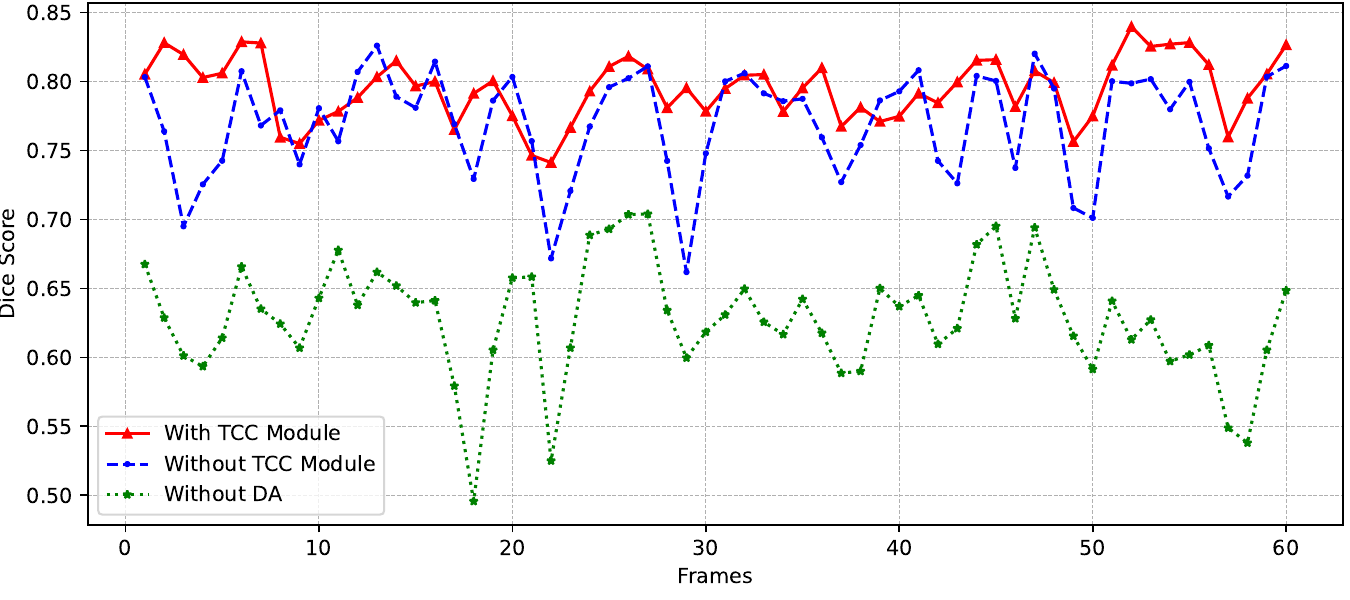}
  \caption{Dice score of the segmentation result for each frame in a video example. 
  The x-axis represents the frame indexes in the video, and the y-axis is the corresponding Dice score.}
  \end{center}
\label{Fig:tcc_statics}
\end{figure}

\noindent\textbf{Ablation study of TCC.}
We explore the effects of two loss functions ($\mathcal{L}_{tc}$ (Eq.~\ref{eq:ltc}) and $\mathcal{L}_{adv}$) in TCC in Table~\ref{tab:tcc}.
In this ablation study, we use SCGM as the baseline model, which has been ablated. 
%
We can see that both $\mathcal{L}_{tc}$ and $\mathcal{L}_{adv}$ can benefit the model, and using two losses can achieve the best performance. For the visualisation of the effectiveness of the TCC module, figure \textcolor{red}{5} illustrates the segmentation result generated by our framework with the TCC module is able to present more consistent performance (marked by the red line) in a video. The results without the TCC module or disabling the domain adaptation perform worse on segmentation consistency.

\noindent \textbf{Effect of the types of attention.} As shown in Table~\ref{tab:abla_attn}, we compare different node attention methods. The results show that inter-intra attention achieves the best performance in our datasets, which indicates the relations between different domains are important to improve the performance.

\noindent \textbf{TCC can learn temporal information.} 
Figure~\textcolor{red}{5} shows the Dice score for each frame in a video example. 
Compared to results without using TCC, our method produces better results with enhanced temporal consistency, showing the effectiveness of the TCC module in learning temporal information. 
\vspace{-5pt}
\section{Conclusion and Limitation}
\vspace{-3pt}
In this paper, we introduced a newly collected CardiacUDA dataset and a novel GraphEcho method for cardiac structure segmentation from echocardiogram videos. 
Our GraphEcho involved two innovative modules, the Spatial-wise Cross-domain Graph Matching (SCGM) and the Temporal Cycle Consistency (TCC) module. 
These two modules are motivated by the fact that the structure of different cardiac structures is similar across different patients and domains and the cyclical consistency property of echocardiogram videos. 
Our approach enables improved UDA segmentation results by effectively aligning global and local features from both source and target domains, thereby preserving both inter-class differences and intra-class similarities.
Experimental results showed that our GraphEcho outperforms existing state-of-the-art UDA segmentation methods. 
%
In our future work, we will explore how to represent objects with complex contours in other medical domains with more efficient representation and conducted the graph-based method on more complicated scenarios such as CT and MRI in future work.

\section*{Acknowledgement}
This work was supported by a research grant from the Beijing Institute of Collaborative Innovation (BICI) under collaboration with the Hong Kong University of Science and Technology under Grant HCIC-004.

{\small
\bibliographystyle{ieee_fullname}
\bibliography{egbib}

\begin{thebibliography}{10}\itemsep=-1pt

\bibitem{chen2020unsupervised}
Cheng Chen, Qi Dou, Hao Chen, Jing Qin, and Pheng~Ann Heng.
\newblock Unsupervised bidirectional cross-modality adaptation via deeply synergistic image and feature alignment for medical image segmentation.
\newblock {\em IEEE TMI}, 39(7):2494--2505, 2020.

\bibitem{chen2020graph}
Liqun Chen, Zhe Gan, Yu Cheng, Linjie Li, Lawrence Carin, and Jingjing Liu.
\newblock Graph optimal transport for cross-domain alignment.
\newblock In {\em International Conference on Machine Learning}, pages 1542--1553. PMLR, 2020.

\bibitem{chen2017rethinking}
Liang-Chieh Chen, George Papandreou, Florian Schroff, and Hartwig Adam.
\newblock Rethinking atrous convolution for semantic image segmentation.
\newblock {\em arXiv preprint arXiv:1706.05587}, 2017.

\bibitem{chen2021semi}
Xiaokang Chen, Yuhui Yuan, Gang Zeng, and Jingdong Wang.
\newblock Semi-supervised semantic segmentation with cross pseudo supervision.
\newblock In {\em Proceedings of the IEEE/CVF Conference on Computer Vision and Pattern Recognition}, pages 2613--2622, 2021.

\bibitem{chen2020self}
Yining Chen, Colin Wei, Ananya Kumar, and Tengyu Ma.
\newblock Self-training avoids using spurious features under domain shift.
\newblock {\em Advances in Neural Information Processing Systems}, 33:21061--21071, 2020.

\bibitem{cuturi2013sinkhorn}
Marco Cuturi.
\newblock Sinkhorn distances: Lightspeed computation of optimal transport.
\newblock {\em Advances in neural information processing systems}, 26, 2013.

\bibitem{dai2022cyclical}
Weihang Dai, Xiaomeng Li, Xinpeng Ding, and Kwang-Ting Cheng.
\newblock Cyclical self-supervision for semi-supervised ejection fraction prediction from echocardiogram videos.
\newblock {\em arXiv preprint arXiv:2210.11291}, 2022.

\bibitem{ding2022exploring}
Xinpeng Ding and Xiaomeng Li.
\newblock Exploring segment-level semantics for online phase recognition from surgical videos.
\newblock {\em IEEE Transactions on Medical Imaging}, 41(11):3309--3319, 2022.

\bibitem{ding2021support}
Xinpeng Ding, Nannan Wang, Shiwei Zhang, De Cheng, Xiaomeng Li, Ziyuan Huang, Mingqian Tang, and Xinbo Gao.
\newblock Support-set based cross-supervision for video grounding.
\newblock In {\em Proceedings of the IEEE/CVF International Conference on Computer Vision}, pages 11573--11582, 2021.

\bibitem{ding2018graph}
Zhengming Ding, Sheng Li, Ming Shao, and Yun Fu.
\newblock Graph adaptive knowledge transfer for unsupervised domain adaptation.
\newblock In {\em Proceedings of the European Conference on Computer Vision (ECCV)}, pages 37--52, 2018.

\bibitem{douglas2011accf}
Pamela~S Douglas, Mario~J Garcia, David~E Haines, Wyman~W Lai, Warren~J Manning, Ayan~R Patel, Michael~H Picard, Donna~M Polk, Michael Ragosta, R~Parker Ward, et~al.
\newblock 2011 appropriate use criteria for echocardiography: a report of the american college of cardiology foundation appropriate use criteria task force.
\newblock {\em Journal of the American College of Cardiology}, 57(9):1126--1166, 2011.

\bibitem{fletcher2021machine}
Andrew~J Fletcher, Winok Lapidaire, and Paul Leeson.
\newblock Machine learning augmented echocardiography for diastolic function assessment.
\newblock {\em Frontiers in Cardiovascular Medicine}, page 879, 2021.

\bibitem{ganin2016domain}
Yaroslav Ganin, Evgeniya Ustinova, Hana Ajakan, Pascal Germain, Hugo Larochelle, Fran{\c{c}}ois Laviolette, Mario Marchand, and Victor Lempitsky.
\newblock Domain-adversarial training of neural networks.
\newblock {\em The journal of machine learning research}, 17(1):2096--2030, 2016.

\bibitem{gong2019dlow}
Rui Gong, Wen Li, Yuhua Chen, and Luc~Van Gool.
\newblock Dlow: Domain flow for adaptation and generalization.
\newblock In {\em Proceedings of the IEEE/CVF conference on computer vision and pattern recognition}, pages 2477--2486, 2019.

\bibitem{goodfellow2020generative}
Ian Goodfellow, Jean Pouget-Abadie, Mehdi Mirza, Bing Xu, David Warde-Farley, Sherjil Ozair, Aaron Courville, and Yoshua Bengio.
\newblock Generative adversarial networks.
\newblock {\em Communications of the ACM}, 63(11):139--144, 2020.

\bibitem{guan2021domain}
Dayan Guan, Jiaxing Huang, Aoran Xiao, and Shijian Lu.
\newblock Domain adaptive video segmentation via temporal consistency regularization.
\newblock In {\em Proceedings of the IEEE/CVF International Conference on Computer Vision}, pages 8053--8064, 2021.

\bibitem{han2022vision}
Kai Han, Yunhe Wang, Jianyuan Guo, Yehui Tang, and Enhua Wu.
\newblock Vision gnn: An image is worth graph of nodes.
\newblock {\em arXiv preprint arXiv:2206.00272}, 2022.

\bibitem{han2021deep}
Xiaoting Han, Lei Qi, Qian Yu, Ziqi Zhou, Yefeng Zheng, Yinghuan Shi, and Yang Gao.
\newblock Deep symmetric adaptation network for cross-modality medical image segmentation.
\newblock {\em IEEE TMI}, 41(1):121--132, 2021.

\bibitem{huang2022category}
Jiaxing Huang, Dayan Guan, Aoran Xiao, Shijian Lu, and Ling Shao.
\newblock Category contrast for unsupervised domain adaptation in visual tasks.
\newblock In {\em Proceedings of the IEEE/CVF Conference on Computer Vision and Pattern Recognition}, pages 1203--1214, 2022.

\bibitem{hughes2021deep}
J~Weston Hughes, Neal Yuan, Bryan He, Jiahong Ouyang, Joseph Ebinger, Patrick Botting, Jasper Lee, John Theurer, James~E Tooley, Koen Nieman, et~al.
\newblock Deep learning evaluation of biomarkers from echocardiogram videos.
\newblock {\em EBioMedicine}, 73:103613, 2021.

\bibitem{kang2020pixel}
Guoliang Kang, Yunchao Wei, Yi Yang, Yueting Zhuang, and Alexander Hauptmann.
\newblock Pixel-level cycle association: A new perspective for domain adaptive semantic segmentation.
\newblock {\em Advances in Neural Information Processing Systems}, 33:3569--3580, 2020.

\bibitem{leclerc2019deep}
Sarah Leclerc, Erik Smistad, Joao Pedrosa, Andreas {\O}stvik, Frederic Cervenansky, Florian Espinosa, Torvald Espeland, Erik Andreas~Rye Berg, Pierre-Marc Jodoin, Thomas Grenier, et~al.
\newblock Deep learning for segmentation using an open large-scale dataset in 2d echocardiography.
\newblock {\em IEEE transactions on medical imaging}, 38(9):2198--2210, 2019.

\bibitem{li2020content}
Guangrui Li, Guoliang Kang, Wu Liu, Yunchao Wei, and Yi Yang.
\newblock Content-consistent matching for domain adaptive semantic segmentation.
\newblock In {\em European conference on computer vision}, pages 440--456. Springer, 2020.

\bibitem{li2022class}
Ruihuang Li, Shuai Li, Chenhang He, Yabin Zhang, Xu Jia, and Lei Zhang.
\newblock Class-balanced pixel-level self-labeling for domain adaptive semantic segmentation.
\newblock In {\em Proceedings of the IEEE/CVF Conference on Computer Vision and Pattern Recognition}, pages 11593--11603, 2022.

\bibitem{li2022sigma}
Wuyang Li, Xinyu Liu, and Yixuan Yuan.
\newblock Sigma: Semantic-complete graph matching for domain adaptive object detection.
\newblock In {\em Proceedings of the IEEE/CVF Conference on Computer Vision and Pattern Recognition}, pages 5291--5300, 2022.

\bibitem{liu2021cycle}
Hong Liu, Jianmin Wang, and Mingsheng Long.
\newblock Cycle self-training for domain adaptation.
\newblock {\em Advances in Neural Information Processing Systems}, 34:22968--22981, 2021.

\bibitem{ma2019gcan}
Xinhong Ma, Tianzhu Zhang, and Changsheng Xu.
\newblock Gcan: Graph convolutional adversarial network for unsupervised domain adaptation.
\newblock In {\em Proceedings of the IEEE/CVF Conference on Computer Vision and Pattern Recognition}, pages 8266--8276, 2019.

\bibitem{mei2020instance}
Ke Mei, Chuang Zhu, Jiaqi Zou, and Shanghang Zhang.
\newblock Instance adaptive self-training for unsupervised domain adaptation.
\newblock In {\em European conference on computer vision}, pages 415--430. Springer, 2020.

\bibitem{melas2021pixmatch}
Luke Melas-Kyriazi and Arjun~K Manrai.
\newblock Pixmatch: Unsupervised domain adaptation via pixelwise consistency training.
\newblock In {\em Proceedings of the IEEE/CVF Conference on Computer Vision and Pattern Recognition}, pages 12435--12445, 2021.

\bibitem{milletari2016v}
Fausto Milletari, Nassir Navab, and Seyed-Ahmad Ahmadi.
\newblock V-net: Fully convolutional neural networks for volumetric medical image segmentation.
\newblock In {\em 2016 fourth international conference on 3D vision (3DV)}, pages 565--571. IEEE, 2016.

\bibitem{oord2018representation}
Aaron van~den Oord, Yazhe Li, and Oriol Vinyals.
\newblock Representation learning with contrastive predictive coding.
\newblock {\em arXiv preprint arXiv:1807.03748}, 2018.

\bibitem{ouyang2020video}
David Ouyang, Bryan He, Amirata Ghorbani, Neal Yuan, Joseph Ebinger, Curtis~P Langlotz, Paul~A Heidenreich, Robert~A Harrington, David~H Liang, Euan~A Ashley, et~al.
\newblock Video-based ai for beat-to-beat assessment of cardiac function.
\newblock {\em Nature}, 580(7802):252--256, 2020.

\bibitem{papolos2016us}
Alexander Papolos, Jagat Narula, Chirag Bavishi, Farooq~A Chaudhry, and Partho~P Sengupta.
\newblock Us hospital use of echocardiography: insights from the nationwide inpatient sample.
\newblock {\em Journal of the American College of Cardiology}, 67(5):502--511, 2016.

\bibitem{tsai2018learning}
Yi-Hsuan Tsai, Wei-Chih Hung, Samuel Schulter, Kihyuk Sohn, Ming-Hsuan Yang, and Manmohan Chandraker.
\newblock Learning to adapt structured output space for semantic segmentation.
\newblock In {\em Proceedings of the IEEE conference on computer vision and pattern recognition}, pages 7472--7481, 2018.

\bibitem{vaswani2017attention}
Ashish Vaswani, Noam Shazeer, Niki Parmar, Jakob Uszkoreit, Llion Jones, Aidan~N Gomez, {\L}ukasz Kaiser, and Illia Polosukhin.
\newblock Attention is all you need.
\newblock {\em Advances in neural information processing systems}, 30, 2017.

\bibitem{vu2019advent}
Tuan-Hung Vu, Himalaya Jain, Maxime Bucher, Matthieu Cord, and Patrick P{\'e}rez.
\newblock Advent: Adversarial entropy minimization for domain adaptation in semantic segmentation.
\newblock In {\em Proceedings of the IEEE/CVF Conference on Computer Vision and Pattern Recognition}, pages 2517--2526, 2019.

\bibitem{wang2020classes}
Haoran Wang, Tong Shen, Wei Zhang, Ling-Yu Duan, and Tao Mei.
\newblock Classes matter: A fine-grained adversarial approach to cross-domain semantic segmentation.
\newblock In {\em European conference on computer vision}, pages 642--659. Springer, 2020.

\bibitem{wang2022semi}
Yuchao Wang, Haochen Wang, Yujun Shen, Jingjing Fei, Wei Li, Guoqiang Jin, Liwei Wu, Rui Zhao, and Xinyi Le.
\newblock Semi-supervised semantic segmentation using unreliable pseudo-labels.
\newblock In {\em Proceedings of the IEEE/CVF Conference on Computer Vision and Pattern Recognition}, pages 4248--4257, 2022.

\bibitem{wu2021unsupervised}
Fuping Wu and Xiahai Zhuang.
\newblock Unsupervised domain adaptation with variational approximation for cardiac segmentation.
\newblock {\em IEEE Transactions on Medical Imaging}, 40(12):3555--3567, 2021.

\bibitem{xie2022towards}
Binhui Xie, Longhui Yuan, Shuang Li, Chi~Harold Liu, and Xinjing Cheng.
\newblock Towards fewer annotations: Active learning via region impurity and prediction uncertainty for domain adaptive semantic segmentation.
\newblock In {\em Proceedings of the IEEE/CVF Conference on Computer Vision and Pattern Recognition}, pages 8068--8078, 2022.

\bibitem{xie2022unsupervised}
Qingsong Xie, Yuexiang Li, Nanjun He, Munan Ning, Kai Ma, Guoxing Wang, Yong Lian, and Yefeng Zheng.
\newblock Unsupervised domain adaptation for medical image segmentation by disentanglement learning and self-training.
\newblock {\em IEEE Transactions on Medical Imaging}, 2022.

\bibitem{yang2022recurring}
Jiewen Yang, Xingbo Dong, Liujun Liu, Chao Zhang, Jiajun Shen, and Dahai Yu.
\newblock Recurring the transformer for video action recognition.
\newblock In {\em Proceedings of the IEEE/CVF Conference on Computer Vision and Pattern Recognition}, pages 14063--14073, 2022.

\bibitem{yang2020fda}
Yanchao Yang and Stefano Soatto.
\newblock Fda: Fourier domain adaptation for semantic segmentation.
\newblock In {\em Proceedings of the IEEE/CVF Conference on Computer Vision and Pattern Recognition}, pages 4085--4095, 2020.

\bibitem{yao2022enhancing}
Huifeng Yao, Xiaowei Hu, and Xiaomeng Li.
\newblock Enhancing pseudo label quality for semi-supervised domain-generalized medical image segmentation.
\newblock In {\em Proceedings of the AAAI Conference on Artificial Intelligence}, pages 3099--3107, 2022.

\bibitem{zhong2021pixel}
Yuanyi Zhong, Bodi Yuan, Hong Wu, Zhiqiang Yuan, Jian Peng, and Yu-Xiong Wang.
\newblock Pixel contrastive-consistent semi-supervised semantic segmentation.
\newblock In {\em Proceedings of the IEEE/CVF International Conference on Computer Vision}, pages 7273--7282, 2021.

\bibitem{zhu2021cross}
Ronghang Zhu, Xiaodong Jiang, Jiasen Lu, and Sheng Li.
\newblock Cross-domain graph convolutions for adversarial unsupervised domain adaptation.
\newblock {\em IEEE Transactions on Neural Networks and Learning Systems}, 2021.

\bibitem{zou2019confidence}
Yang Zou, Zhiding Yu, Xiaofeng Liu, BVK Kumar, and Jinsong Wang.
\newblock Confidence regularized self-training.
\newblock In {\em Proceedings of the IEEE/CVF International Conference on Computer Vision}, pages 5982--5991, 2019.

\end{thebibliography}
}

\section*{Appendix A: Algorithm Pipeline}
Based on the GraphEcho that we have presented in Section \textcolor{red}{3}, the SGCM and TCC module can be formulated as Algorithm~\ref{algorithm:SGCM} and  Algorithm~\ref{algorithm:TCC}. Note that all the superscript $s$ and $t$ of all variables represent both source and target domains. For instance, $\mathbf{x}^{s,t}$ indicate the input from both source $\mathbf{x}^{s}$ and target $\mathbf{x}^{t}$ domains.

\section*{Appendix B: Visualization for sequences of echocardiogram videos}
\begin{figure*}[t]
  \begin{center}
  \includegraphics[width=0.999\linewidth]{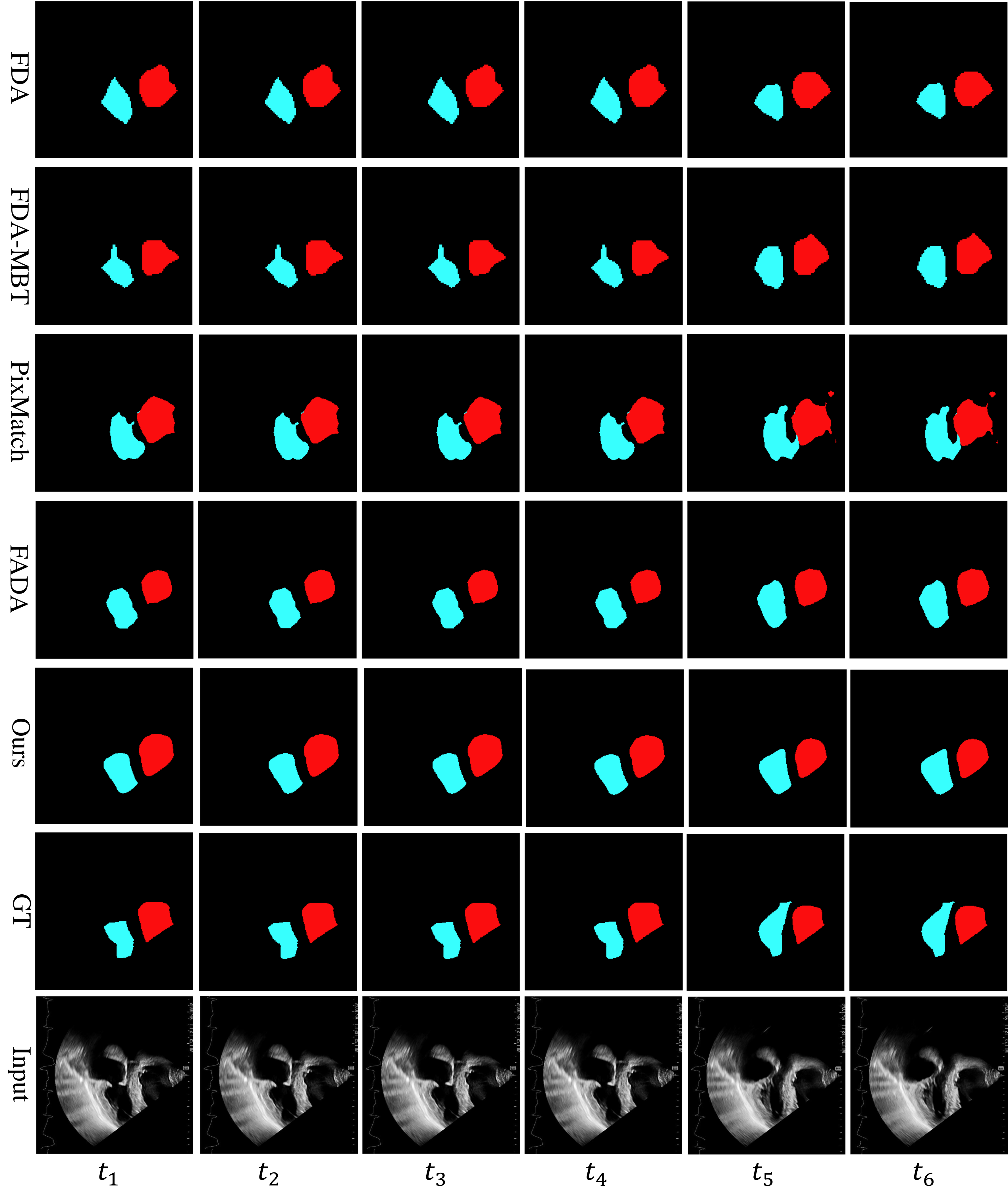}
  \caption{We visualize a video sequence of parasternal left ventricle long axis view to show the segmentation results (GT denotes the ground truth segmentation result). Red and cyan indicate refer to the segmentation regions for the right Atrium (RA) and left atrium (LV), respectively. The $\{t1,...,t6\}$ denotes the frame order of a video sequence.}
  \label{Fig:video_visulization_view1}
  \end{center}
\end{figure*}

\begin{figure*}[t]
  \begin{center}
  \includegraphics[width=0.999\linewidth]{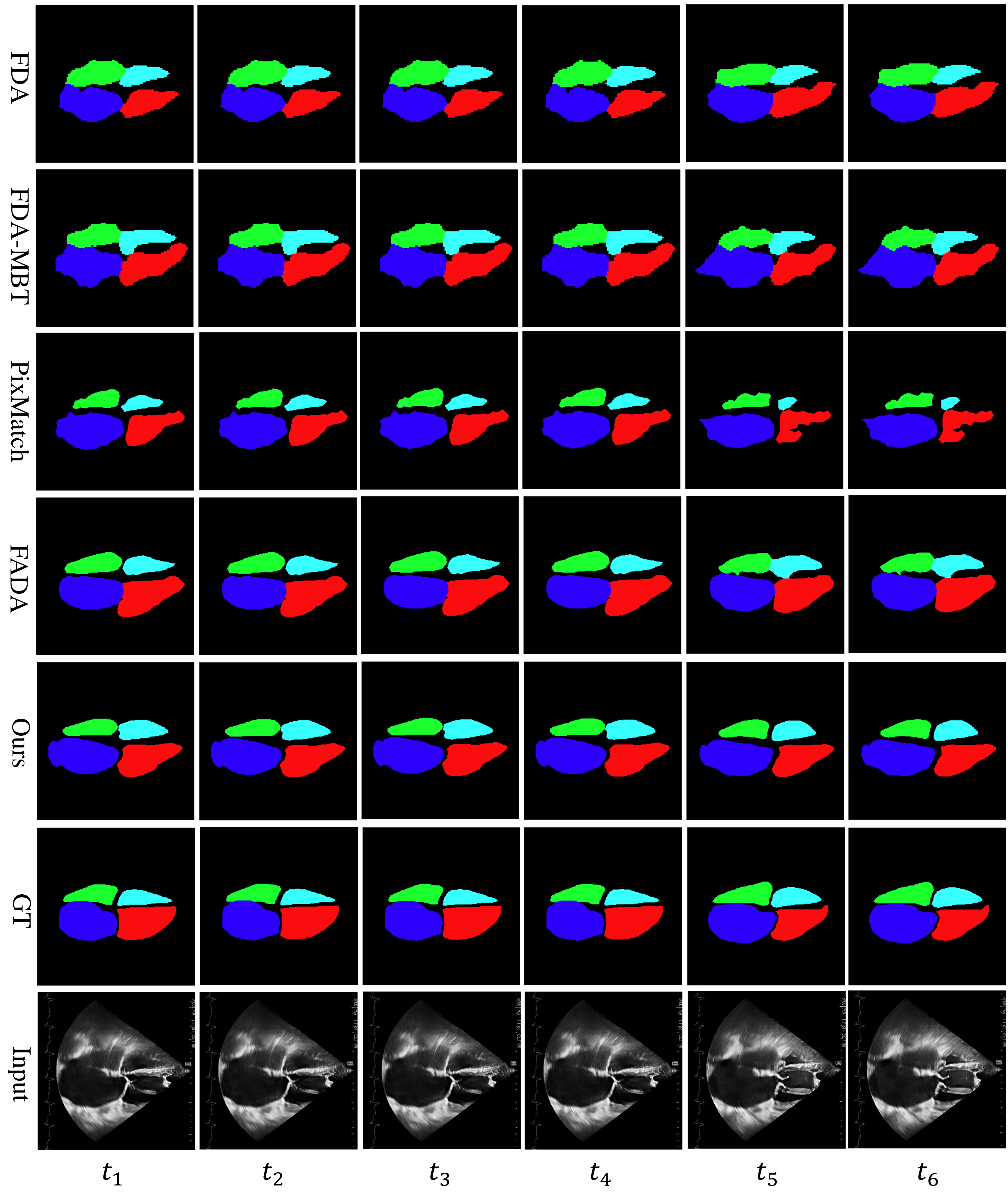}
  \caption{We visualize a video sequence of an apical four-chamber heart view to show the segmentation results. Red, green, blue, and cyan indicate refer to the segmentation regions for the right Atrium (RA), left ventricle (LV), right ventricle (RV), and left atrium (LV), respectively. The $\{t1,...,t6\}$ denotes the frame order of a video sequence.}
  \label{Fig:video_visulization_view2}
  \end{center}
\end{figure*}

In this supplementary, we provide more visualization results (see figure~\ref{Fig:video_visulization_view1} and~\ref{Fig:video_visulization_view2}) for the sequences of echocardiogram videos.

\begin{algorithm*}[t!]
  \renewcommand{\algorithmicrequire}{\textbf{Input:}}
  \renewcommand{\algorithmicensure}{\textbf{Output:}}
  \caption{Spatial-wise Cross-domain Graph Matching Module (SGCM)}
  \begin{algorithmic}[1]

    \ENSURE ~~\\
    $\mathcal{L}_{SCGM}$ : The overall loss of the SGCM;\\
    $\mathcal{L}_{cls}$ : The classification loss of the SGCM;\\
    $\mathcal{L}_{mat}$ : The graph matching loss of the SGCM;\\
    $\mathcal{L}_{seg}$ : Supervised segmentation loss;\\
    $\mathcal{L}_{bce}$ : Binary cross-entropy loss;\\
    $\mathcal{L}_{dice}$ : Dice loss~\cite{milletari2016v}.\\
    \vspace{12pt}
    
    \REQUIRE ~~\\
    $\mathbf{x}^{s,t}$ : Input video frames from source and target domains;\\
    $\mathbf{y}^{s}$ : The ground truth annotation of the source domain;\\
    $m$ : The number of the sampling feature;\\
    $C$ : The total classes number of segmentation region;\\
    $\alpha, \beta$ : The classification loss weight for source and target domains;\\
    $\text{Encoder($\cdot$)}$ : The Parameter shared feature extractor;\\
    $\text{Decoder($\cdot$)}$ : The Parameter shared decoder for generating the segmentation result;\\
    $\mathbb{I}(\cdot)$ : Indicator function.
    \vspace{12pt}

    \STATE $\mathbf{f}^{s,t}$ $\leftarrow$ $\text{Encoder}(\mathbf{x}^{s,t})$
    \STATE $\hat{\mathbf{y}}^{s,t} \leftarrow \text{Decoder}(\mathbf{f}^{s,t})$
    \STATE $\mathcal{L}_{seg} \leftarrow \mathcal{L}_{bce}(\hat{\mathbf{y}}^s, {\mathbf{y}}^s) +  \mathcal{L}_{dice}(\hat{\mathbf{y}}^s, {\mathbf{y}}^s)$
    \vspace{12pt}

    \textit{\textbf{(1).Node Sampling:}}
    \FOR{$i=1$ \textbf{to} \textbf{in} $C$}
      \STATE $\{\mathbf{f}_i^{s,t}\}$ $\leftarrow$ Get different chamber region $\{\mathbf{f}_i^{s,t}\}$ according to $\mathbf{y}^{s}$ and $\hat{\mathbf{y}}^{t}$.
      \STATE $\mathbf{v}^{s,t}$ $\leftarrow$ Uniformly sample $m$ feature vectors from $\{\mathbf{f}_i^{s,t}\}$ as the node embedding $\mathbf{v}^{s,t}$.
    \ENDFOR
    \vspace{12pt}

    \textit{\textbf{(2).Node Classification:}}
    \STATE $\bar{\mathbf{v}}^{s,t} \leftarrow Attention(concat(\mathbf{v}^{s},\mathbf{v}^{t}))$
    \STATE $\mathcal{L}_{cls} \leftarrow -\alpha  \mathbf{y} log(h(\bar{\mathbf{v}}^{s})) - \beta  \hat{\mathbf{y}} log(h(\bar{\mathbf{v}}^{t}))$
    \vspace{12pt}

    \textit{\textbf{(3).Graph Matching:}}
    \STATE $g^{s,t}$ $\leftarrow$ Add the learned matrix as the edge connections $\mathbf{e}^{s,t}$ to $\mathbf{v}^{s,t}$ and constructed semantic graph $g^{s,t}$.
    \STATE $\mathbf{A} \leftarrow $ Obtain adjacency matrix $\mathbf{A}$ from $g^{s}$ and $g^{t}$.
    \STATE $\vec{\mathbf{A}} \leftarrow Sinkhorn(\mathbf{A})$: Obtain transport cost matrix of graphs among chambers.
    \STATE Minimize the transport distance of $p$-th row and $q$-th column element on $\vec{\mathbf{A}}$.
    $\mathcal{L}_{mat} \leftarrow  \sum_{p,q} \; \mathbb{I}(\mathbf{y}_p^s=\hat{\mathbf{y}}_q^t) \cdot min(\vec{\mathbf{A}}(p,q)) + \mathbb{I}(\mathbf{y}_p^s\neq \hat{\mathbf{y}}_q^t) \cdot max(\vec{\mathbf{A}}(p,q))$. 
    \vspace{12pt}

    \textit{\textbf{Overall Loss:}}
    \STATE $\mathcal{L}_{SCGM}=\mathcal{L}_{cls}+\mathcal{L}_{mat}.$

  \end{algorithmic}
  \label{algorithm:SGCM}
  \end{algorithm*}
%
%
\begin{algorithm*}[h]
  \renewcommand{\algorithmicrequire}{\textbf{Input:}}
  \renewcommand{\algorithmicensure}{\textbf{Output:}}
  \caption{Temporal-wise Cycle Consistency Module (TCC)}
  \begin{algorithmic}[1]
    \ENSURE ~~\\
    $\mathcal{L}_{tc}^{s,t}$ : The temporal consistency loss for the source and target domains; \\
    $\mathcal{L}_{TCC}$ : The overall loss of temporal consistency. \\
    
    \REQUIRE ~~\\
    $\mathbf{X}^{s,t}$ : Input video from source and target domains.\\
    $\mathcal{L}_{adv}$ : The adversarial methods~\cite{ganin2016domain} to eliminate the domain gap with global feature alignment;\\
    $N$ : The number of frames in $\mathbf{X}^{s,t}$;\\
    $\mathbf{h}_t$ : The hidden state, and the $h_0$ is learned parameters with all zero in initial state;\\
    $\mathbf{w}_{gcn},\mathbf{b}_{gcn}$ :  The graph convolution weight and bias;\\
    $\sigma$ : The activation function;\\
    $FFN$ :  Feed forward network;\\
    $\text{Encoder($\cdot$)}$ : The Parameter shared feature extractor;\\
    $\text{RGCC($\cdot$)}$ : Recursive graph convolutional cell.\\
    \vspace{12pt}

    \textit{\textbf{(1).Temporal Graph Node Construction:}}
    \STATE $\{ \mathbf{f}_i^{s,t} \}_{i=1}^N \leftarrow \text{Encoder}(\mathbf{X}^{s,t})$, Where $\mathbf{f}_i$ is the feature of the $i$-th frame.
    \STATE $\{ \ddot{\mathbf{v}}_i^{s,t} \}_{i=1}^N \leftarrow$ Apply average pooling and node sampling (see algorithm~\ref{algorithm:SGCM}.(1)) to feature map $\{ \mathbf{f}_i^{s,t} \}_{i=1}^N$.
    \vspace{12pt}

    \textit{\textbf{(2).Recursive Graph Convolutional Cell (RGCC):}}
    \FOR{$t=0$ \textbf{to} $N$}
      \STATE $\{ \ddot{\mathbf{v}}^{s,t}_t, \ddot{\mathbf{e}}^{s,t}_t \} \leftarrow $ Find $K$ nearest neighbors on the hidden state $\mathbf{h}_t$ for each node at $\ddot{\mathbf{v}}^{s,t}_t$.
      \STATE $\mathbf{h}_{t+1}^{s,t} \leftarrow \sigma \mathbf{w}_{gcn}(\ddot{\mathbf{v}}^{s,t}_i, \ddot{\mathbf{e}}^{s,t}_i) + \mathbf{b}_{gcn}$
    \ENDFOR
    \STATE $\mathbf{o}^{s,t} \leftarrow FFN (\mathbf{h}_N^{s,t})$
    \vspace{12pt}

    \textit{\textbf{(3).Temporal consistency loss (source domain as an example):}}
    \STATE $\mathcal{L}_{tc}^s \leftarrow -\sum_{ \{ \mathbf{o}^s_{k}, \mathbf{o}^s_{+}\} \in \mathcal{P}^s} log\frac{\text{exp}(\mathbf{o}^s_{k} \cdot \mathbf{o}^s_{+})}{\sum_{ \mathbf{o}^s_{-} \in \mathcal{B}} \text{exp}(\mathbf{o}^s_{k} \cdot \mathbf{o}^s_{-})}$ \\
    where $\mathcal{P}^s$ is the set of positive pairs, $\mathbf{o}^s_{k}$ and $\mathbf{o}^s_{+}$ are representations that randomly sampled from a video $\mathbf{X}^s$ as positive pairs. For the negative samples, we maintain a memory bank $\mathcal{B}$ consisting of representations of clips sampled from different videos.
    \vspace{12pt}
    
    \textit{\textbf{Overall Loss:}}
    \STATE $\mathcal{L}_{TCC} = \mathcal{L}_{tc}^{s,t}+\mathcal{L}_{adv}$
    
  \end{algorithmic}
  \label{algorithm:TCC}
  \end{algorithm*}

\end{document}